\documentclass[10pt,twocolumn,letterpaper]{article}

\usepackage{cvpr}
\usepackage{times}
\usepackage{epsfig}
\usepackage{graphicx}
\usepackage{amsmath}
\usepackage{amssymb}

\usepackage[pagebackref=true,breaklinks=true,letterpaper=true,colorlinks,bookmarks=false]{hyperref}

\newcommand{\argmax}{\operatornamewithlimits{argmax}}

\cvprfinalcopy 

\def\cvprPaperID{1055} 

\ifcvprfinal\fi
\begin{document}

\title{Designing Deep Convolutional Neural Networks for Continuous \\Object Orientation Estimation}

\author{Kota Hara, Raviteja Vemulapalli and Rama Chellappa\\
Center for Automation Research, UMIACS, University of Maryland, College Park\\
}

\maketitle

\begin{abstract}
Deep Convolutional Neural Networks (DCNN) have been proven to be effective for various computer vision problems. In this work, we demonstrate its effectiveness on a continuous object orientation estimation task, which requires prediction of $0^{\circ}$ to $360^{\circ}$ degrees orientation of the objects. We do so by proposing and comparing three continuous orientation prediction approaches designed for the DCNNs. The first two approaches work by representing an orientation as a point on a unit circle and minimizing either L2 loss or angular difference loss. The third method works by first converting the continuous orientation estimation task into a set of discrete orientation estimation tasks and then converting the discrete orientation outputs back to the continuous orientation using a mean-shift algorithm. By evaluating on a vehicle orientation estimation task and a pedestrian orientation estimation task, we demonstrate that the discretization-based approach not only works better than the other two approaches but also achieves state-of-the-art performance. We also demonstrate that finding an appropriate feature representation is critical to achieve a good performance when adapting a DCNN trained for an image recognition task.
\end{abstract}

\section{Introduction}


The effectiveness of the Deep Convolutional Neural Networks (DCNNs) has been demonstrated for various computer vision tasks such as image classification \cite{He2015a,Krizhevsky2012,Szegedy2015,Simonyan2015a}, object detection \cite{Girshick2014,Girshick2015,Ren2015}, semantic-segmentation \cite{Long2014,Chen2014a,Noh2015,Pinheiro2014}, human body joint localization \cite{Toshev2014,Tompson2014,Chen2014b}, face recognition \cite{Taigman2014} and so on. Due to the large number of network parameters that need to be trained, DCNNs require a significant number of training samples. For tasks where sufficient number of training samples are not available, a DCNN trained on a large dataset for a different task is tuned to the current task by making necessary modifications to the network and retraining it with the available training data \cite{Yosinski2014a}. This transfer learning technique has been proven to be effective for tasks such as fine-grained recognition \cite{Sharif2014,Gui2015,Azizpour2015}, object detection \cite{Girshick2014,Sermanet2014a}, object classification \cite{Zeiler2014}, attribute detection \cite{Sharif2014,Azizpour2015} and so on.

One of the tasks for which a large number of training samples are not available is the continuous object orientation estimation problem where the goal is to predict the continuous orientation of objects in the range $0^{\circ}$ to $360^{\circ}$. The orientations are important properties of objects such as pedestrians and cars and precise estimation of the orientations allows better understanding of the scenes essential for applications such as autonomous driving and surveillance. Since in general it is difficult to annotate the orientations of objects without a proper equipment, the number of training samples in existing datasets for orientation estimation tasks is limited. Thus, it would be interesting to see if and how it would be possible to achieve good performance by adapting a DCNN trained on a large object recognition dataset \cite{Russakovsky2014} to the orientation estimation task. 

The first consideration is the representation used for prediction.
When a DCNN is trained for an image classification task, the layers inside the network gradually transform the raw pixel information to more and more abstract representation suitable for the image classification task. Specifically, to achieve good classification ability, the representations at later layers have more invariance against shift/rotation/scale changes while maintaining a good discriminative power between different object classes. On the other hand, the orientation estimation task requires representation which can capture image differences caused by orientation changes of the objects in the same class. Thus, it is important to thoroughly evaluate the suitability of representations from different layers of a DCNN trained on the image classification task for the object orientation estimation task.

The second consideration is the design of the orientation prediction unit.
For the continuous orientation estimation task, the network has to predict the angular value, which is in a non-Euclidean space, prohibiting the direct use of a typical L2 loss function. To handle this problem, we propose three different approaches. The first approach represents an orientation as a 2D point on a unit circle, then trains the network using the L2 loss. In test time, the network's output, a 2D point not necessarily on a unit circle, is converted back to the angular value by $\mathrm{atan2}$ function. Our second approach also uses a-point-on-a-unit-circle representation, however, instead of the L2 loss, it minimizes a loss defined directly on the angular difference. Our third approach, which is significantly different from the first two approaches, is based on the idea of converting the continuous orientation estimation task into a set of discrete orientation estimation tasks and addressing each discrete orientation estimation task by a standard softmax function. In test time, the discrete orientation outputs are converted back to the continuous orientation using a mean-shift algorithm. The discretized orientations are determined such that all the discretized orientations are uniformly distributed in the output circular space. The mean-shift algorithm for the circular space is carried out to find the most plausible orientation while taking into account the softmax probability for each discrete orientation. 

We conduct experiments on car orientation estimation and pedestrian orientation estimation tasks. We observe that the approach based on discretization and mean-shift algorithm outperforms the other two approaches with a large margin. We also find that the final performance significantly varies with the feature map used for orientation estimation. We believe that the findings from the experiments reported here can be beneficial for other object classes as well. 



The paper is organized as follows. Section \ref{sec:related_work} discusses related works. Section \ref{sec:methods} presents three proposed approaches. Section \ref{sec:experiments} shows experimental results and conclusions are given in section \ref{sec:conclusions}.

\section{Related Work}\label{sec:related_work}

Object orientation estimation problem has been gaining more and more attention due to its practical importance. Several works treat the continuous orientation estimation problem as a multi-class classification problem by discretizing the orientations. In \cite{Ozuysal2009}, a three-step approach is proposed where a bounding box containing the object is fist predicted, then orientation is estimated based on image features inside the predicted bounding box, and finally a classifier tuned for the predicted orientation is applied to check the existence of the object. \cite{Ghodrati2014} address the orientation estimation task using Fisher encoding and convolutional neural network-based features. \cite{Bakry2014} learns a visual manifold which captures large variations in object appearances and proposes a method which can untangle such a visual manifold into a view-invariant category representation and a category-invariant pose representation. 

Some approaches address the task as continuous prediction in order to avoid undesirable approximation error caused by discretization. In \cite{He2015}, a joint object detection and orientation estimation algorithm based on structural SVM is proposed. In order to effectively optimize a nonconvex objective function, a cascaded discrete-continuous inference algorithm is introduced. In \cite{Hara2014,Hara2016}, a regression forest trained with a multi-way node splitting algorithm is proposed. As an image descriptor, HOG features are used. \cite{Teney2014} introduces a representation along with a similarity measure of 2D appearance based on distributions of low-level, fine-grained image features. For continuous prediction, an interpolation based approach is applied. A neural network-based model called Auto-masking Neural Network (ANN) for joint object detection and view-point estimation is introduced in \cite{Yang2014}. The key component of ANN is a mask layer which produces a mask passing only the important part of the image region in order to allow only these regions to be used for the final prediction. Although both our method and ANN are neural network-based methods, the overall network architectures and the focus of the work are significantly different.

Several works consider learning a suitable representation for the orientation estimation task. In \cite{Torki2011}, an embedded representation that reflects the local features and their spatial arrangement as well as enforces supervised manifold constraints on the data is proposed. Then a regression model to estimate the orientation is learned using the proposed representation. Similarly to \cite{Torki2011}, \cite{Fenzi2013,Fenzi2014} learn a representation using spectral clustering and then train a single regression for each cluster while enforcing geometric constraints. \cite{Fenzi2015} formulates the task as a MAP inference task, where the likelihood function is
composed of a generative term based on the prediction error
generated by the ensemble of Fisher regressors as well
as a discriminative term based on SVM classifiers. 


\cite{Xiang2014} introduces PASCAL3D+ dataset designed for joint object detection and pose estimation. Continuous annotations of azimuth and elevation for 12 object categories are provided. The average number of instances per category is approximately 3,000. The performance is evaluated based on Average Viewpoint Precision (AVP) which takes into account both the detection accuracy and view-point estimation accuracy. Since the focus of this work is the orientation estimation, we employ the EPFL Multi-view Car Dataset \cite{Ozuysal2009} and the TUD Multiview Pedestrian Dataset \cite{Andriluka2010} specifically designed to evaluate the orientation prediction.

Despite the availability of continuous ground-truth view point information, majority of works \cite{Xiang2014,Tulsiani2015a,Massa2016,Pepik2012,Ghodrati2014,Su2015} using PASCA3D+ dataset predict discrete poses and evaluate the performance based on the discretized poses. \cite{Tulsiani2015a} proposes a method for joint view-point estimation and key point prediction based on CNN. It works by converting the continuous pose estimation task into discrete view point classification task. \cite{Su2015} proposes to augment training data for their CNN model
by synthetic images. The view point prediction is cast as a fine-grained (360 classes for each angle) discretized view point classification problem.

\section{Method}\label{sec:methods}
Throughout this work we assume that a single object, viewed roughly from the side, is at the center of the image and the orientation of the object is represented by a value ranging from $0^{\circ}$ to $360^{\circ}$. Before feeding to the network, we first resize the input image to a canonical size and then subtracted the dataset mean. The network then processes the input image by applying a series of transformations, followed by an orientation prediction unit producing the final estimates. In this section, we present each of the proposed orientation prediction units in details. All the prediction units are trained by back propagation. 


\subsection{Orientation Estimation}

\subsubsection{Approach 1}
We first represent orientation angles as points on a unit circle by $v = ( \cos{\theta}, \sin{\theta} )$. We then train a standard regression layer with a Huber loss function, also known as smooth L1 loss function. The Huber loss is used to improve the robustness against outliers, however, a standard L2 or L1 loss can also be used when appropriate. During testing, predicted 2D point coordinate $v = (x,y)$ is converted to the orientation angle by $\theta = \mathrm{atan2}(y,x)$. A potential issue in this approach is that the Huber loss function, as well as L2 or L1 loss functions, consider not only the angular differences but also the radial differences that are not directly related to the orientation.

\subsubsection{Approach 2}
As in approach 1, we represent orientation angles as points on a unit circle and train a regression function, however, we use a loss function which focuses only on the angular differences:
\begin{equation}
L(v_g, v ) = 1-\cos(\theta) = 1 - \frac{ v_g \cdot v }{|v_g||v|} = 1 - \frac{ x_g x + y_g y }{ \sqrt{ x^2 + y^2 } }
\end{equation}
where $v_g = (x_g, y_g )$ is the ground-truth. Note that $|v_g| = 1$ by definition. The derivative of $L$ with respect to $x$ is computed as
\begin{equation}
\frac{\partial L}{\partial x} = \frac{(x_g x + y_g y) \frac{x}{\sqrt{x^2 + y^2}} -  x_g \sqrt{x^2 + y^2} }{x^2 + y^2}
\end{equation}
We compute $\frac{\partial L}{\partial y}$ similarly. These derivatives allow us to train the network parameters by back propagation. As in approach 1, during testing, the predicted 2D point coordinates are converted to orientation angles by the atan2 function.

A potential issue in this approach is that the derivatives approaches 0 when angular difference becomes close to $180^{\circ}$, making the optimization more challenging. 

\subsubsection{Approach 3}
\begin{figure*}[tbh]
\begin{center}
\includegraphics[width=6.0in]{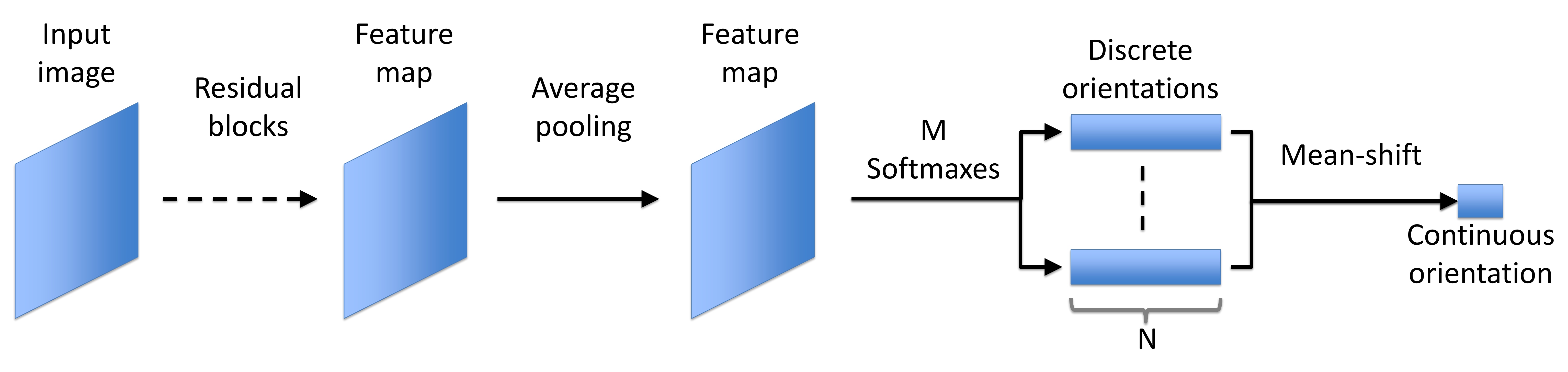}
\caption{The network arthitecture for the discretization based approach. \label{fig:network2}}
\end{center}
\end{figure*}

\begin{figure}[tbh]
\begin{center}
\includegraphics[width=1.7in]{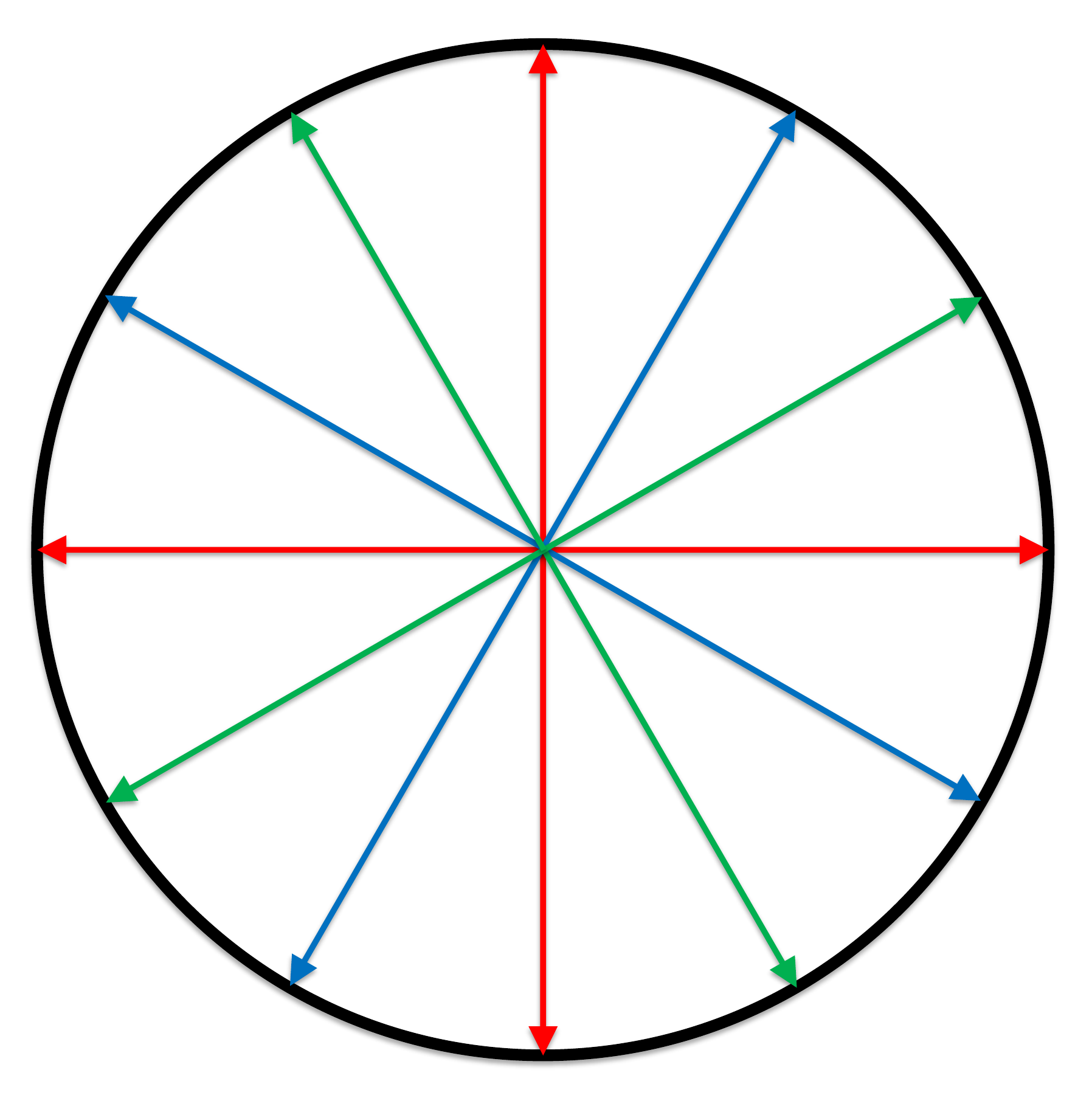}
\caption{The example discretization constructed by $N=4, M=3$. $N$ Orientations depicted in the same color are used for one of the $M$ classification task. \label{fig:cropped_discretization}}
\end{center}
\end{figure}

We propose an approach based on discretization. The network architecture illustrating this approach is presented in Fig.~\ref{fig:network2}. We first discretize the 0-360 range into $N$ unique orientations which are $G=360/N^{\circ}$ degree apart and convert the continuous prediction task into an $N$-class classification task. Each training sample is assigned one of the $N$ class labels based on its orientation's proximity to the discretized orientations. In order to alleviate the loss of information introduced by the discretization, we construct $M$ classification tasks by having a different starting orientation for each discretization. The $M$ starting orientations equally divide $G$ degree. Formally, the discrete orientations for the $m$-th classification task are $\{m \times G/M + k \times G\}_{k=0, \dots, N-1}$, where $m=0, \dots, M-1$. The example discretization with $N=4, M=3$ is depicted in Fig.~\ref{fig:cropped_discretization}. As an orientation estimation unit, we thus have $M$ independent $N$-way softmax classification layers, which are trained jointly.

During testing, we compute softmax probabilities for each of the $M$ classification tasks separately. Consequently, we obtain probabilistic votes at all of the $M \times N$ unique orientations. We then define a probability density function using weighted kernel density estimation adopted for the circular space. We use the von-Mises distribution as a kernel function. The von-Mises kernel is defined as
\begin{equation}\label{eq:von_Mises}
k_\nu (\theta)  = \frac{1}{2 \pi I_0(\nu)} \exp{( \nu \cdot \cos( \theta ))}
\end{equation}
where $\nu$ is the concentration parameter and $I_0(\nu)$ is the modified Bessel function of order 0.

Formally, the density at the orientation $\theta$ is given by
\begin{equation}
\hat{p}( \theta ; \nu ) \propto \sum_{i=1}^{M \times N} p_i k_\nu ( \theta - \theta_i).
\end{equation}
where $\theta_i$ is the $i$-th discrete orientation and $p_i$ is the corresponding softmax probability.

Then final prediction is made by finding the orientation with the highest density:
\begin{equation}
\hat{\theta} = \argmax_{\theta} \hat{p}( \theta ; \nu )
\end{equation}

In order to solve the above maximization problem, we use a mean-shift mode seeking algorithm specialized for a circular space proposed in \cite{Hara2016} 


The same level of discretization can be achieved by different combinations of $N$ and $M$. For instance, both $(N,M)=(72,1)$ and $(N,M)=(8,9)$ discretize the orientation into 72 unique orientations, however, we argue that larger $N$ makes the classification task more difficult and confuses the training algorithm since there are smaller differences in appearances among neighboring orientations. Setting $M$ larger than 1 and reducing $N$ could alleviate this problem while maintaining the same level of discretization. This claim is verified through the experiments.

A potential problem of the proposed approach is the loss of information introduced by the discretization step. However, as shown later in the experiment section, the mean-shift algorithm successfully recovers the continuous orientation without the need for further discretization.

\section{Experiments}\label{sec:experiments}
We evaluate the effectiveness of the proposed approaches on the EPFL Multi-view Car Dataset \cite{Ozuysal2009} and TUD Multiview Pedestrian Dataset \cite{Andriluka2010}. Both datasets have continuous orientation annotations available.

\subsection{DCNN}
As an underlying DCNN, we employ the Residual Network \cite{He2015a}  with 101 layers (ResNet-101) pre-trained on ImageNet image classification challenge \cite{Russakovsky2014} with 1000 object categories. The ResNet-101 won the 1st place on various competitions such as ImageNet classification, ImageNet detection, ImageNet localization, COCO detection, and COCO segmentation. Although ResNet-152, which is deeper than ResNet-101, achieves better performance than the ResNet-101, we employ ResNet-101 due to its smaller memory footprint.

The key component of the ResNet is a residual block, which is designed to make the network training easier. The residual block is trained to output the residual with reference to the input to the block. The residual block can be easily constructed by adding the input to the output from the block. ResNet-101 consists of 33 residual blocks. Each residual block contains three convolution layers, each of which is followed by a Batch Normalization layer, a scale Layer and the ReLU layer. 

\subsection{Training details}
Unless otherwise noted, the weights of the existing ResNet-101 layers are fixed to speed up the experiments. The parameters of the orientation prediction unit are trained by Stochastic Gradient Descent (SGD). All the experiments are conducted using the Caffe Deep Learning Framework \cite{Jia2014} on a NVIDIA K40 GPU with 12GB memory. In order to include contextual regions, bounding box annotations are enlarged by a scale factor of 1.2 and 1.1 for EPFL and TUD datasets, respectively. We augment the training data by including the vertically mirrored versions of the samples. 

For all experiments, we apply average pooling with size 3 and stride 1 after the last residual block chosen and then attach the orientation prediction unit. The batch size, momentum and weight decay are set to 32, 0.9 and 0.0005, respectively. Weights of the all the orientation prediction layers are initialized by random numbers generated from the zero-mean Gaussian distribution with $\mathrm{std} = 0.0001$. All biases are initialized to 0.

For the approach 3, we set $M$, the number of starting orientations for discretization, to 9 and $N$ to 8 for all the experiments unless otherwise stated. 

\subsection{EPFL dataset}

The EPFL dataset contains 20 sequences of images of cars captured at a car show where each sequence contains images of the same instance of car captured with various orientations. Each image is given a ground-truth orientation. We use the first 10 sequences as training data and the remaining 10 sequences for testing. As a result, the number of training samples is 2,358 after data augmentation and that of testing samples is 1,120. We use bounding box information which comes with the dataset to crop out the image region to be fed to the network. The performance of the algorithm is measured by Mean Absolute Error (MeanAE) and Median Absolute Error (MeadianAE) in degree following the practice in the literature. Unless otherwise noted, the number of training iterations is 2000 with 0.000001 as a learning rate, followed by the additional 2,000 iterations with 10 times reduced learning rate. 

First we conduct experiments to figure out the most suitable representation for the orientation estimation task by attaching the orientation prediction unit to different residual blocks. For these experiments, we use approach 3. 

\begin{figure}[tbh]
\begin{center}
\includegraphics[width=2.8in]{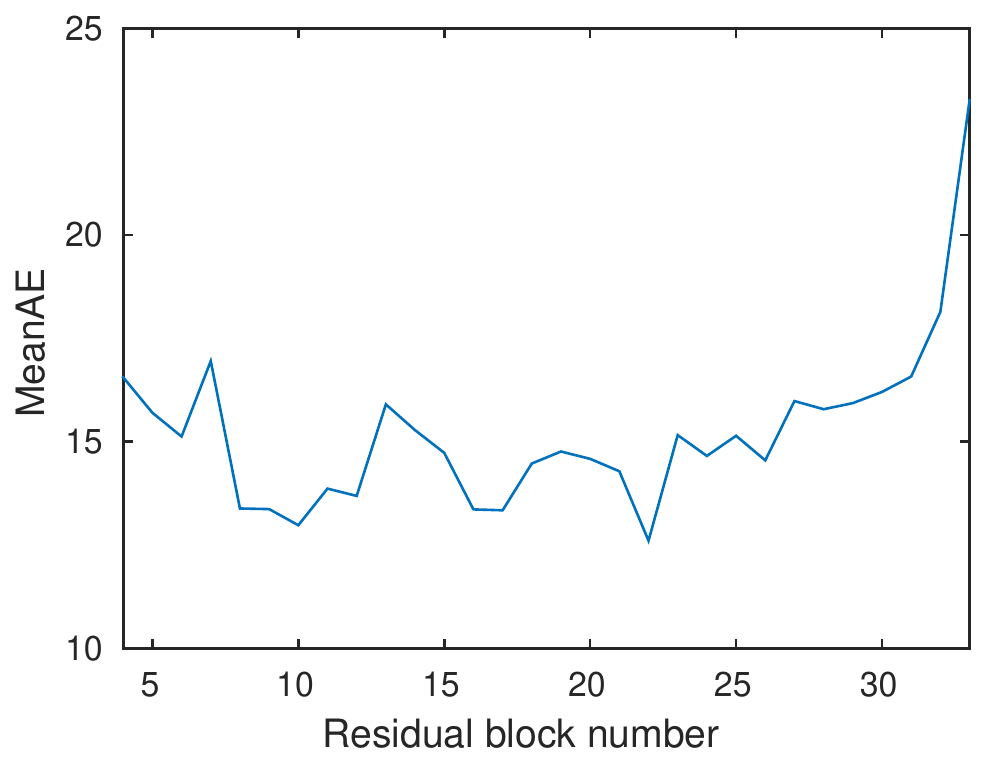}
\caption{The performance on EPFL dataset obtained by using a different residual block to attach the proposed orientation prediction unit. \label{fig:feature_map_EPFL}}
\end{center}
\end{figure}

In Fig.~\ref{fig:feature_map_EPFL}, we show the MeanAE on the EPFL dataset obtained by using different residual blocks. As can be seen, both the earlier and later residual blocks do not provide a suitable representation for the orientatin estimation task. The 22nd residual block produces the best representation among 33 residual blocks for our task. The following experiments are conducted using the 22nd residual block.

We analyze the effect of $M$, the number of starting orientations for discretization, and $N$, the number of unique orientations in each discretization, of the discretization-based approach. The results are summarized in Table.~\ref{tb:effect_of_M}. It is observed that when $N=8$, increasing $M$ leads to better results, however, no significant improvement is observed after $M=9$. When $N=72$, increasing $M$ from 1 to 5 does not lead to better performance. These results indicate that the larger number of the total orientations leads to better performance upto some point. 

When the total number of unique orientations is same, e.g., $(N,M) = (72,1)$ and $(N,M) = (8,9)$, MeanAE is smaller with $(N,M) = (8,9)$ while MedianAE is smaller with $(N,M) = (72,1)$. Since MedianAE is very small with both settings, both of them achieve high accuracy in most of the cases.

\begin{table*}[htb]
\small
\begin{center}
  \caption{MeanAE and MedianAE with different values for $N$ and $M$ on the EPFL dataset.\label{tb:effect_of_M}}
  \begin{tabular}{|c||c|c|c|c|c|c|c|} \hline
   $N$ & \multicolumn{2}{|c|}{72} & \multicolumn{5}{|c|}{8} \\ \hline
   $M$ & 1 & 5 & 1 & 3 & 5 & 9 & 15 \\ \hline
   MeanAE & 13.6 & 13.7 & 19.0 & 13.3 & 13.2 & 12.6 & 12.6 \\ \hline
   MedianAE & 2.7 & 2.8 & 10.8 & 3.0 & 3.0 & 3.0 & 3.1 \\ \hline
   \end{tabular}
\end{center}
\end{table*}


Table \ref{tb:other_approaches_EPFL} shows the performance of the other two approaches based on a-point-on-a-unit-circle representation. For the approach 1, we train the model with the same setting used for the approach 3. For the approach 2, we train the model for 40,000 iterations with a learning rate of 0.0000001 as it appears necessary for convergence. As can be seen, the discretization-based approach significantly outperforms the other two approaches. 

\begin{table}
\small
\begin{center}
  \caption{Comparison among the proposed approaches on the EPFL dataset.\label{tb:other_approaches_EPFL}}
  \begin{tabular}{|c|c|c|} \hline
   Approach & MeanAE & MedianAE \\ \hline \hline
   1 & 22.9 & 11.3 \\ \hline
   2 & 26.7 & 10.2 \\ \hline
   3 & 12.6 & 3.0 \\ \hline   
   \end{tabular}
\end{center}
\end{table}

In Table ~\ref{tb:existing_work}, we present results from the literature and the result of our final model ( approach 3 ). For this comparison, instead of fixing the existing ResNet weights, we fine-tune all the network parameters end-to-end, which reduce the MeanAE by 21.7\%.  As can be seen, our final model advances the state of the art performance. 

The information on whether or not the ground-truth bounding box annotations are used in test time is also included in the table. In methods which do not utilize the ground-truth bounding boxes, an off-the-shelf object detector such as DPM \cite{Felzenszwalb2010} is used to obtain the bounding boxes \cite{Fenzi2015} or the localization and orientation estimation are addressed jointly \cite{He2015,Yang2014,Teney2014,Redondo-cabrera2014,Ozuysal2009}.

\begin{table*}
\small
\begin{center}
  \caption{Comparison with the existing works on the EPFL dataset. The performance is measured in Mean Absolute Error (MeanAE) and Median Absolute Error (MedianAE).\label{tb:existing_work}}
  \begin{tabular}{|c|c|c|c|} \hline
   Methods & MeanAE & MedianAE & Ground-truth Bounding box? \\ \hline \hline
   \textbf{Ours} & \textbf{9.86} & \textbf{3.14} & Yes \\ \hline
   Fenzi et al. \cite{Fenzi2015} & 13.6 & 3.3 & No \\ \hline   
   He et al. \cite{He2015}  & 15.8 & 6.2 & No \\ \hline
   Fenzi and Ostermann \cite{Fenzi2014} & 23.28 & N/A & Yes \\ \hline
   Hara and Chellappa \cite{Hara2016}   & 23.81 & N/A & Yes \\ \hline
   Zhang at al. \cite{Zhang2013} & 24.00 & N/A & Yes \\ \hline
   Yang et al. \cite{Yang2014} & 24.1 & 3.3 & No \\ \hline
   Hara and Chellappa \cite{Hara2014}   & 24.24 & N/A & Yes \\ \hline
   Fenzi et al. \cite{Fenzi2013} & 31.27 & N/A & Yes \\ \hline
   Torki and Elgammal \cite{Torki2011} & 33.98 & 11.3 & Yes \\ \hline
   Teney and Piater \cite{Teney2014}  & 34.7 & 5.2 & No \\ \hline   
   Redondo-Cabrera et al. \cite{Redondo-cabrera2014} & 39.8 & 7 & No \\ \hline
   Ozuysal et al. \cite{Ozuysal2009} & 46.5 & N/A & No \\ \hline
   \end{tabular}
\end{center}
\end{table*}

\begin{figure*}[!htbp]
\centering
\begin{tabular}{cccccccccc}
\includegraphics[width=0.52in]{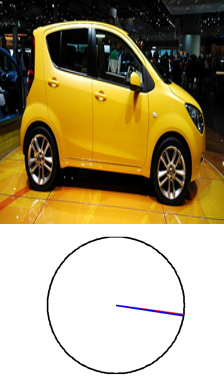} & 	        \includegraphics[width=0.52in]{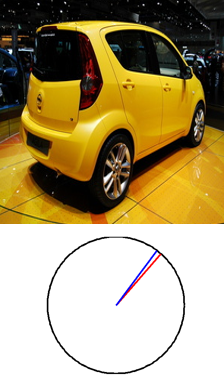} &
\includegraphics[width=0.52in]{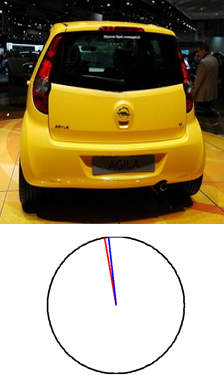} &
\includegraphics[width=0.52in]{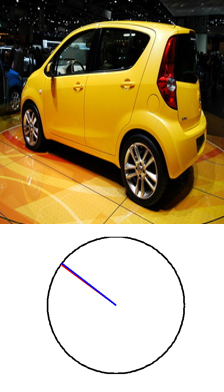} &
\includegraphics[width=0.52in]{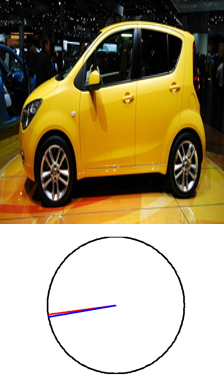} &
\includegraphics[width=0.52in]{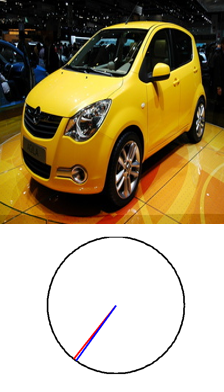} &
\includegraphics[width=0.52in]{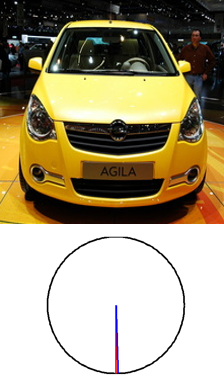} &
\includegraphics[width=0.52in]{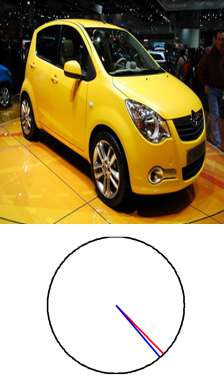} &
\includegraphics[width=0.52in]{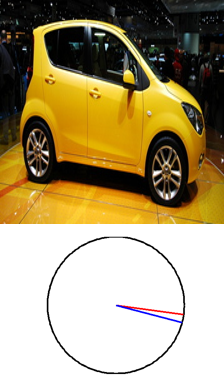} &
\includegraphics[width=0.52in]{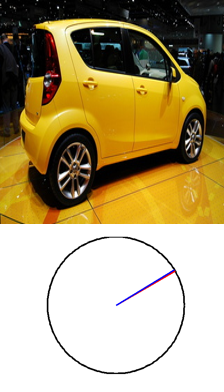}
\\
\includegraphics[width=0.52in]{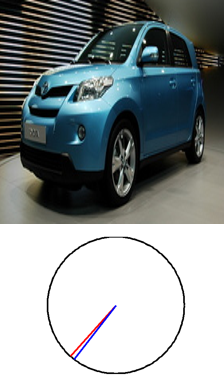} & 	        \includegraphics[width=0.52in]{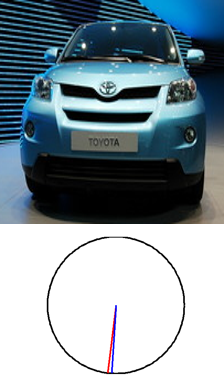} &
\includegraphics[width=0.52in]{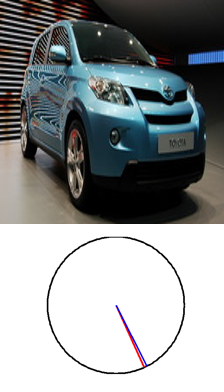} &
\includegraphics[width=0.52in]{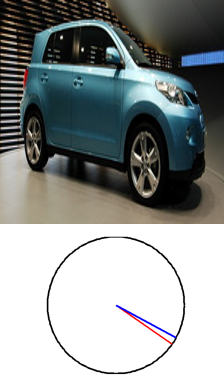} &
\includegraphics[width=0.52in]{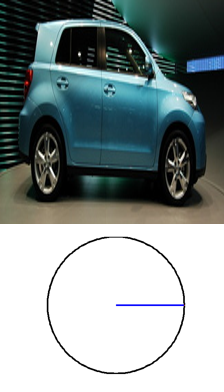} &
\includegraphics[width=0.52in]{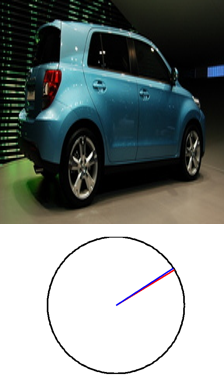} &
\includegraphics[width=0.52in]{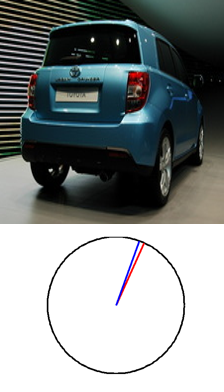} &
\includegraphics[width=0.52in]{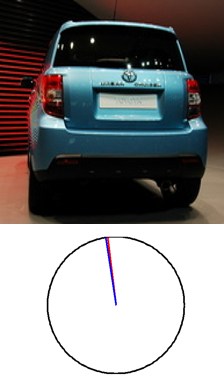} &
\includegraphics[width=0.52in]{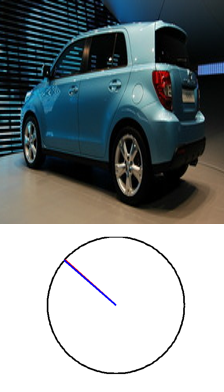} &
\includegraphics[width=0.52in]{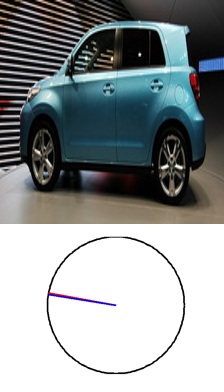}
\\
\includegraphics[width=0.52in]{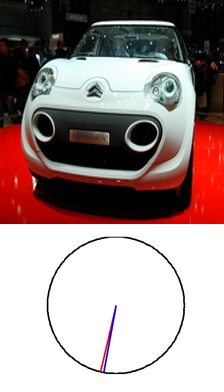} & 	        \includegraphics[width=0.52in]{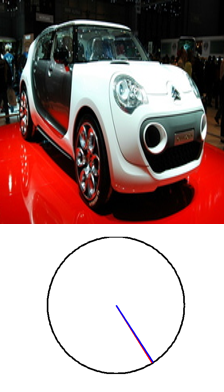} &
\includegraphics[width=0.52in]{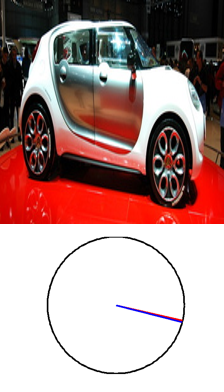} &
\includegraphics[width=0.52in]{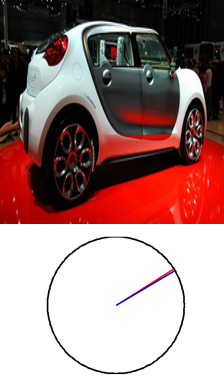} &
\includegraphics[width=0.52in]{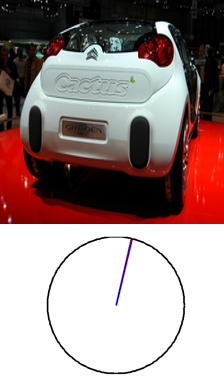} &
\includegraphics[width=0.52in]{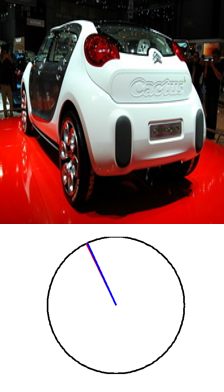} &
\includegraphics[width=0.52in]{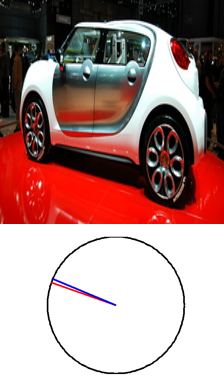} &
\includegraphics[width=0.52in]{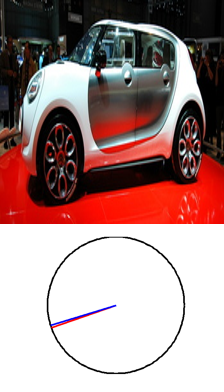} &
\includegraphics[width=0.52in]{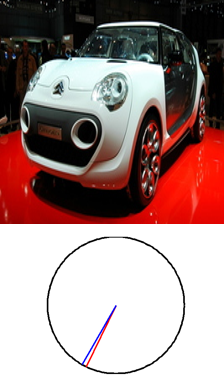} &
\includegraphics[width=0.52in]{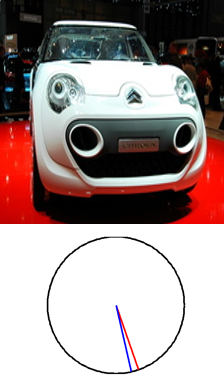}
\\
\includegraphics[width=0.52in]{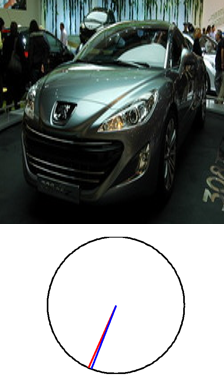} & 	        \includegraphics[width=0.52in]{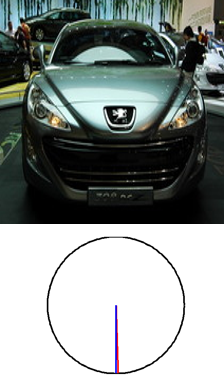} &
\includegraphics[width=0.52in]{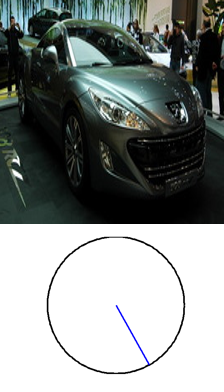} &
\includegraphics[width=0.52in]{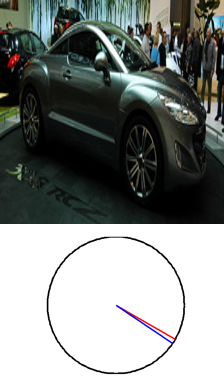} &
\includegraphics[width=0.52in]{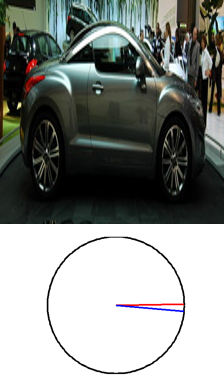} &
\includegraphics[width=0.52in]{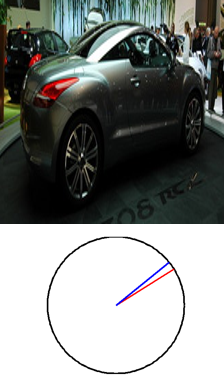} &
\includegraphics[width=0.52in]{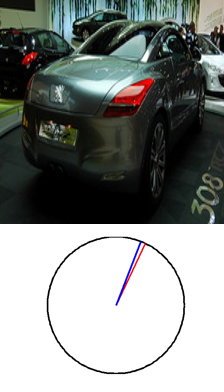} &
\includegraphics[width=0.52in]{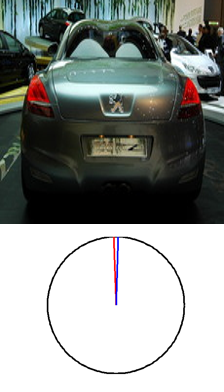} &
\includegraphics[width=0.52in]{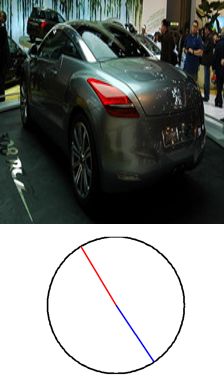} &
\includegraphics[width=0.52in]{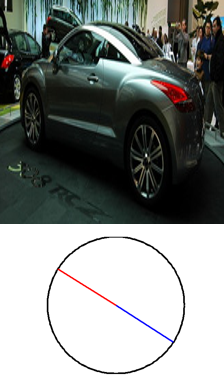}
\\
\includegraphics[width=0.52in]{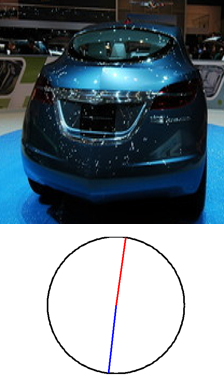} & 	        \includegraphics[width=0.52in]{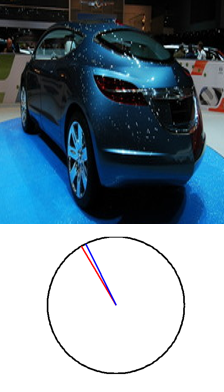} &
\includegraphics[width=0.52in]{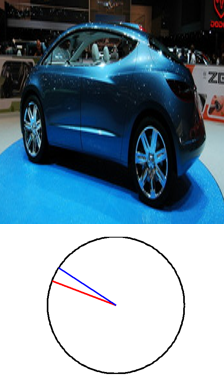} &
\includegraphics[width=0.52in]{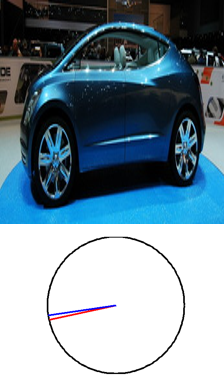} &
\includegraphics[width=0.52in]{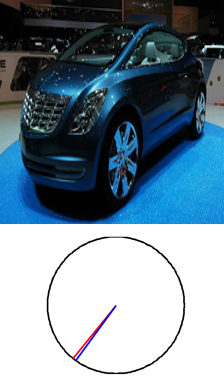} &
\includegraphics[width=0.52in]{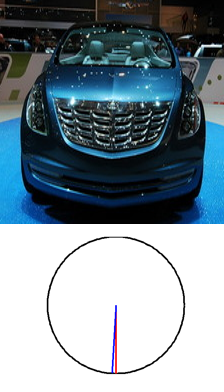} &
\includegraphics[width=0.52in]{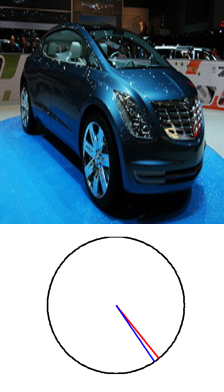} &
\includegraphics[width=0.52in]{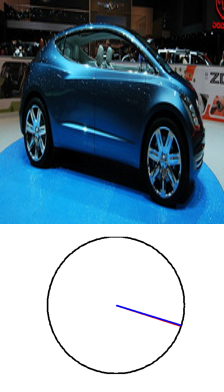} &
\includegraphics[width=0.52in]{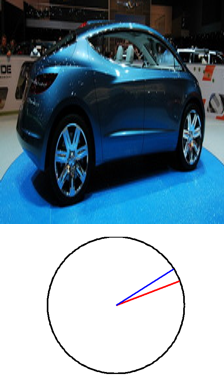} &
\includegraphics[width=0.52in]{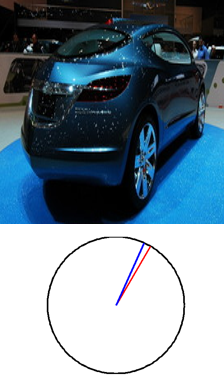}
\\
\end{tabular}
\caption{Representative results obtained by the proposed method ( approach 3, $N=8,M=9$). A ground-truth orientation (red) and predicted orientation (blue) are indicated in a circle. Each row contains 10 example results from a testing sequence. From left to right, images are selected with 10 frames apart, starting from the first frame.}\label{fig:caffe_results_EPFL}
\end{figure*}

Finally, we show representative results in Fig.~\ref{fig:caffe_results_EPFL} with ground-truth bounding boxes overlaid on the images and a ground-truth orientation and predicted orientation indicated in a circle. Note that many of the failure cases are due to the flipping errors ($\approx 180^\circ$) and tend to occur at a specific instance whose front and rear look similar (See the last two examples in the row 4.)

\subsection{TUD dataset}

The TUD dataset consists of 5,228 images of pedestrians with bounding box annotations. Since the original annotations are discrete orientations, we use continuous annotations provided by \cite{Hara2016}. In total, there are 4,732 images for training, 290 for validation and 309 for testing. Note that the size of the dataset is more than two times larger than that of EPFL Multi-view Car Dataset and unlike the EPFL dataset, images are captured in the wild. Since most of the training images are gray scale images and thus not adequate to feed into the DCNN, we convert all the grey scale images into color images by a recently proposed colorization technique \cite{zhang2016}. The performance of the algorithm is measured by Mean Absolute Error (MeanAE), Accuracy-22.5 and Accuracy-45 as in \cite{Hara2016}. Accuracy-22.5 and Accuracy-45 are defined as the ratio of samples whose
predicted orientation is within $22.5^{\circ}$ and $45^{\circ}$ from the ground truth, respectively. For this dataset, the number of training iterations is 10,000 with 0.00001 as a learning rate. 

\begin{figure}[htb]
\begin{center}
\includegraphics[width=2.8in]{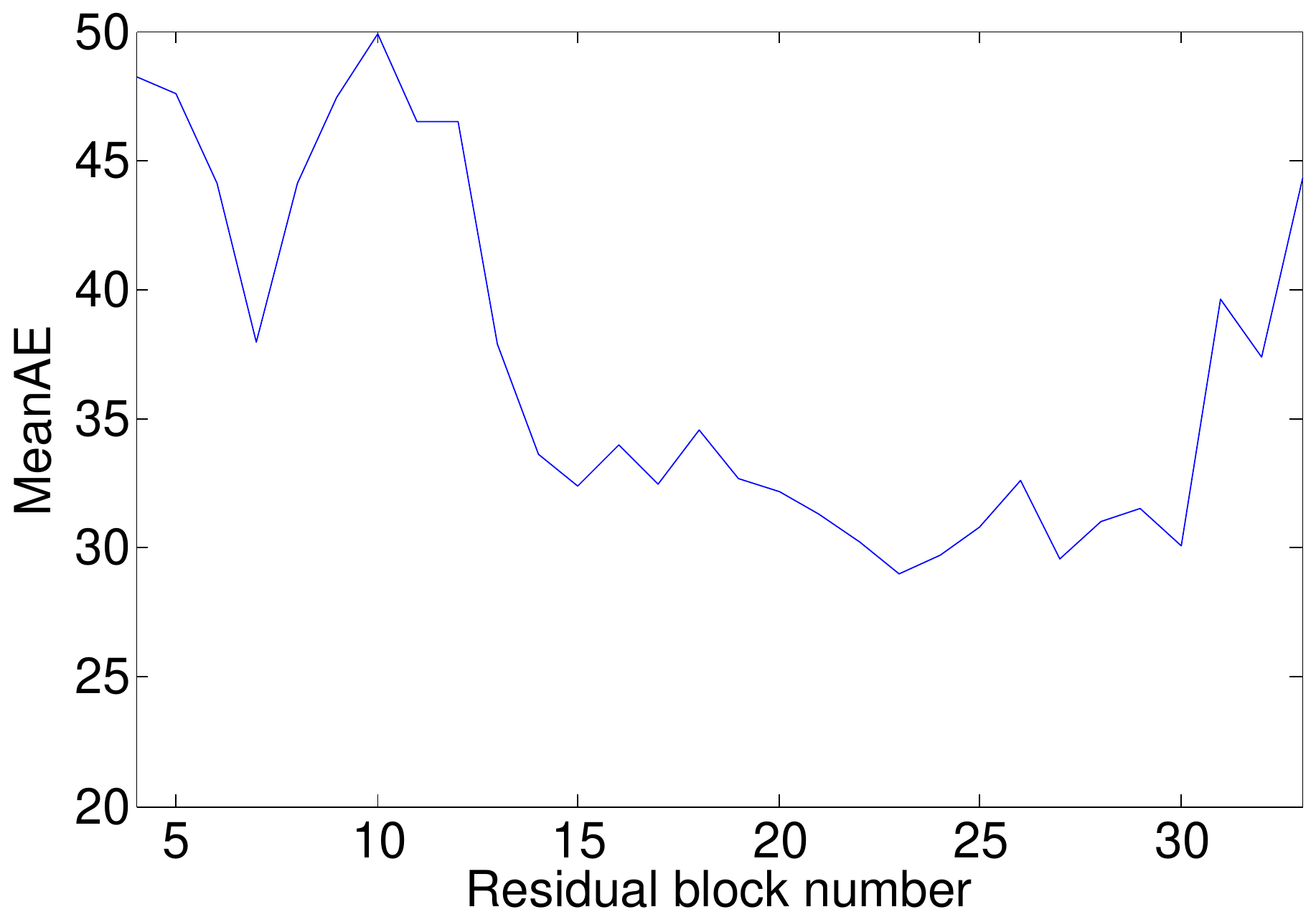}
\caption{The performance on TUD dataset obtained by using a different residual block to attach the proposed orientation prediction unit. \label{fig:feature_map_TUD}}
\end{center}
\end{figure}

In Fig.~\ref{fig:feature_map_TUD}, we show the MeanAE obtained by attaching the orientation estimation unit of the approach 3 to different residual blocks. As is the case with the EPFL dataset, the performance varies significantly depending on the residual block used. Furthermore, the use of a proper representation is more critical on this dataset. Interestingly though, as in the EPFL dataset, the 22nd residual block performs well. Following experiments are thus conducted by using the 22nd residual block. 

In Table.~\ref{tb:effect_of_M_TUD}, we show the effect of $M$ while keeping $N=8$. As can be seen, in general larger $M$ produces better results, however, no significant improvement is observed after $M=9$. In order to evaluate the effect of having multiple non-overlapping discretization, we compare $N=72, M=1$ setting, whose number of discrete angles is same as $N=8, M=9$ setting. As can be seen in the table, the effect of having multiple discretization is prominent.

\begin{table*}[htb]
\small
\begin{center}
  \caption{MeanAE, Accuracy-22.5 and Accuracy-45 with different $M$ on the TUD dataset.\label{tb:effect_of_M_TUD}}
  \begin{tabular}{|c||c|c|c|c|c|c|c|} \hline
   $N$ &  \multicolumn{2}{|c|}{72} & \multicolumn{5}{|c|}{8} \\ \hline
   $M$ & 1 & 5 & 1 & 3 & 5 & 9 & 15 \\ \hline
   MeanAE & 35.4 & 33.5 & 40.0 & 32.7 & 31.1 & 30.2 & 30.9 \\ \hline
   Accuracy-22.5 & 63.1 & 62.5 & 55.0 & 61.5 & 61.5 & 63.1 & 62.5  \\ \hline
   Accuracy-45 & 79.6 & 80.6 & 75.4 & 79.6 & 82.2 & 82.8 & 82.5 \\ \hline
   \end{tabular}
\end{center}
\end{table*}

Table \ref{tb:other_approaches_TUD} shows the performance of all the proposed approaches. For the approach 2, we increase the training iterations to 70,000 as it appears to take more iterations to converge. It is observed again that the approach 3 performs best.

\begin{table}
\small
\begin{center}
  \caption{Comparison among the proposed approaches on the TUD dataset.\label{tb:other_approaches_TUD}}
  \begin{tabular}{|c|c|c|c|} \hline
   Approach & MeanAE & Accuracy-22.5 & Accuracy-45 \\ \hline \hline
   1 & 33.7 & 46.9 & 75.7 \\ \hline
   2 & 34.6 & 44.0 & 70.9 \\ \hline
   3 & 30.2 & 63.1 & 82.8 \\ \hline   
   \end{tabular}
\end{center}
\end{table}

Finally, we train our model by fine-tuning all the layer parameters end-to-end. The result is shown in Table \ref{tb:TUD_results} along with the result of prior art. The end-to-end training reduces the MeanAE by 11.9 \%. Our final model outperforms the state-of-the-art with 23.3 \% reduction in MeanAE. The table contains the performance of human which is significantly better than the algorithms, necessitating further algorithm development.

\begin{figure*}[htb]
\centering
\begin{tabular}{cccccccccc}
\includegraphics[width=0.51in]{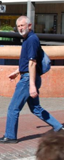} & 	        \includegraphics[width=0.51in]{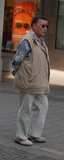} &
\includegraphics[width=0.51in]{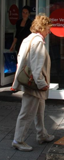} &
\includegraphics[width=0.51in]{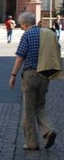} &
\includegraphics[width=0.51in]{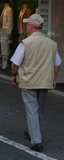} &
\includegraphics[width=0.51in]{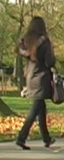} &
\includegraphics[width=0.51in]{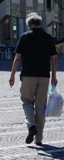} &
\includegraphics[width=0.51in]{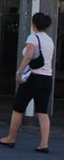} &
\includegraphics[width=0.51in]{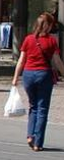} &
\includegraphics[width=0.51in]{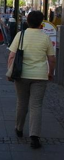}
\\
\includegraphics[width=0.51in]{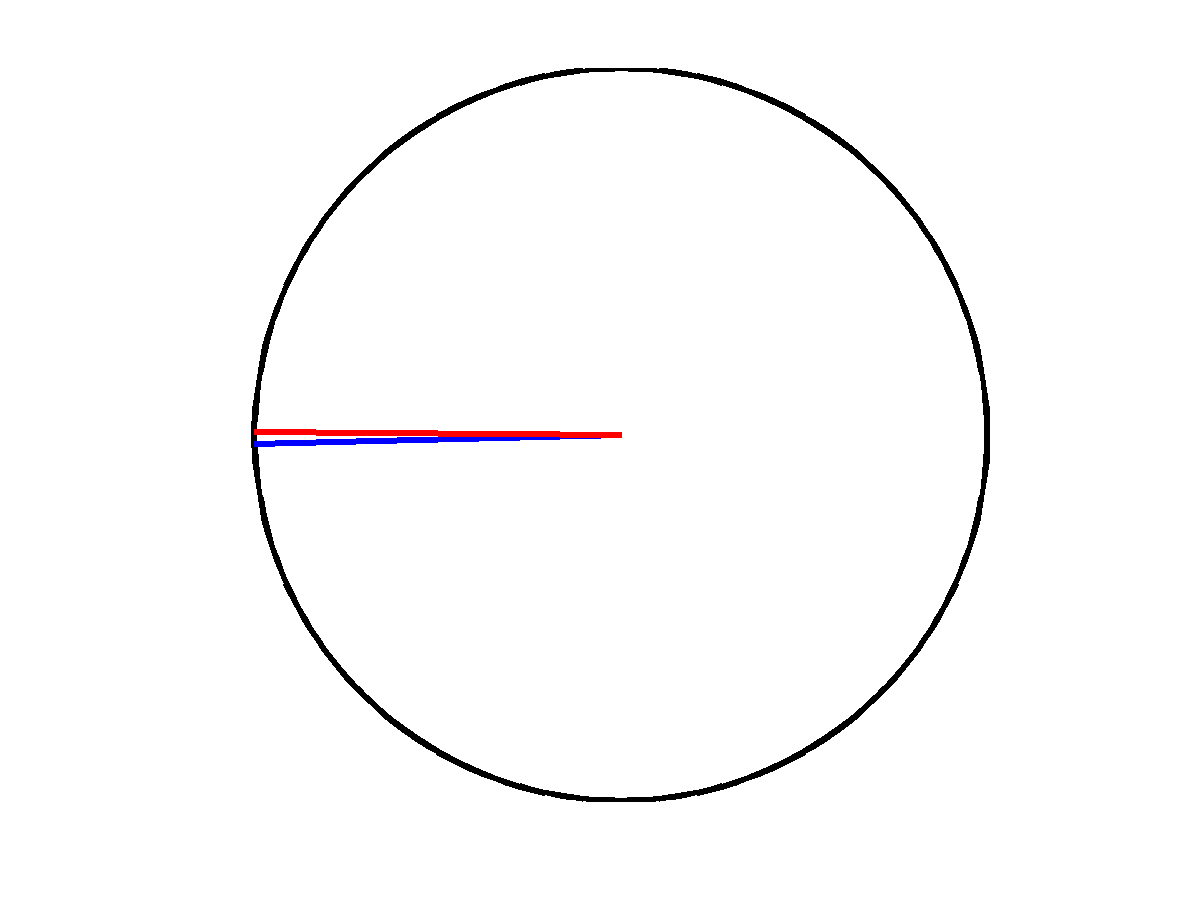} & 	        \includegraphics[width=0.51in]{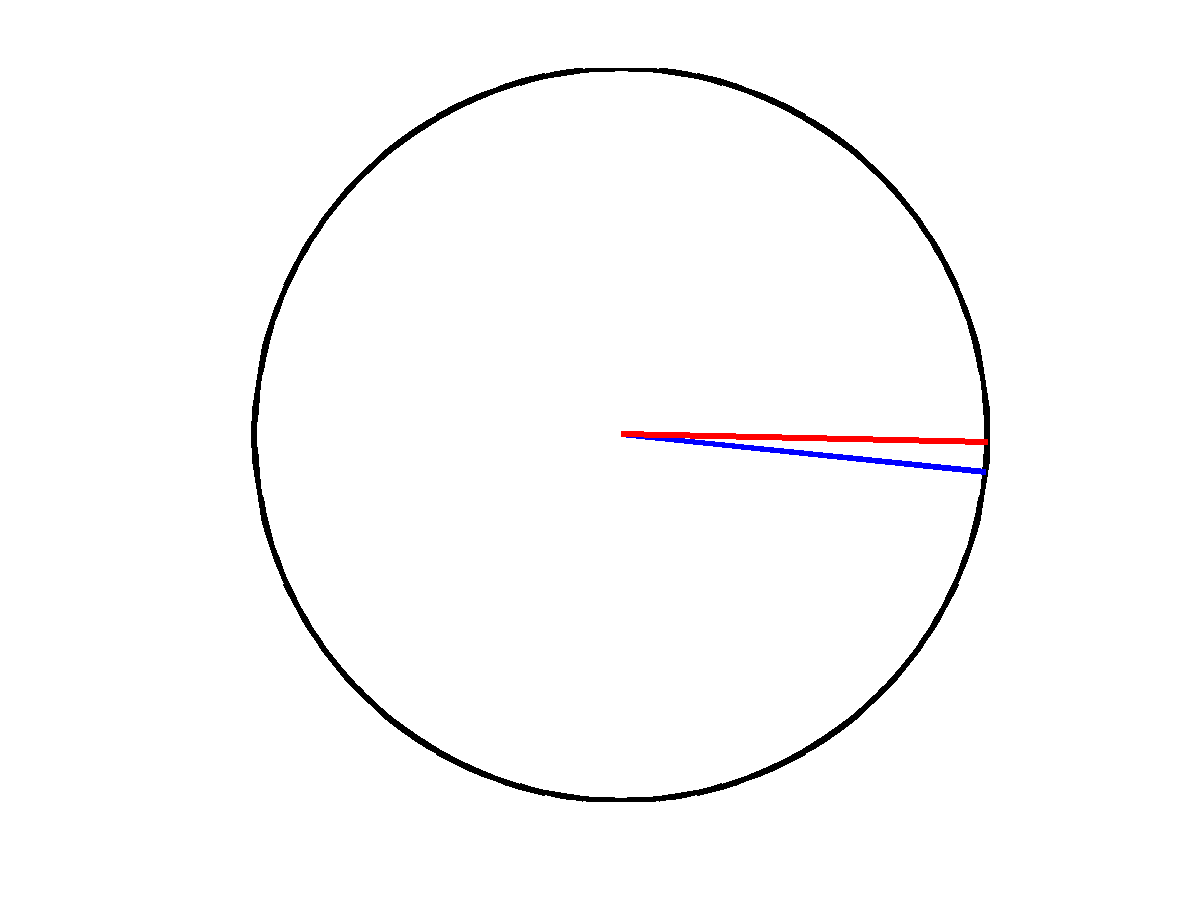} &
\includegraphics[width=0.51in]{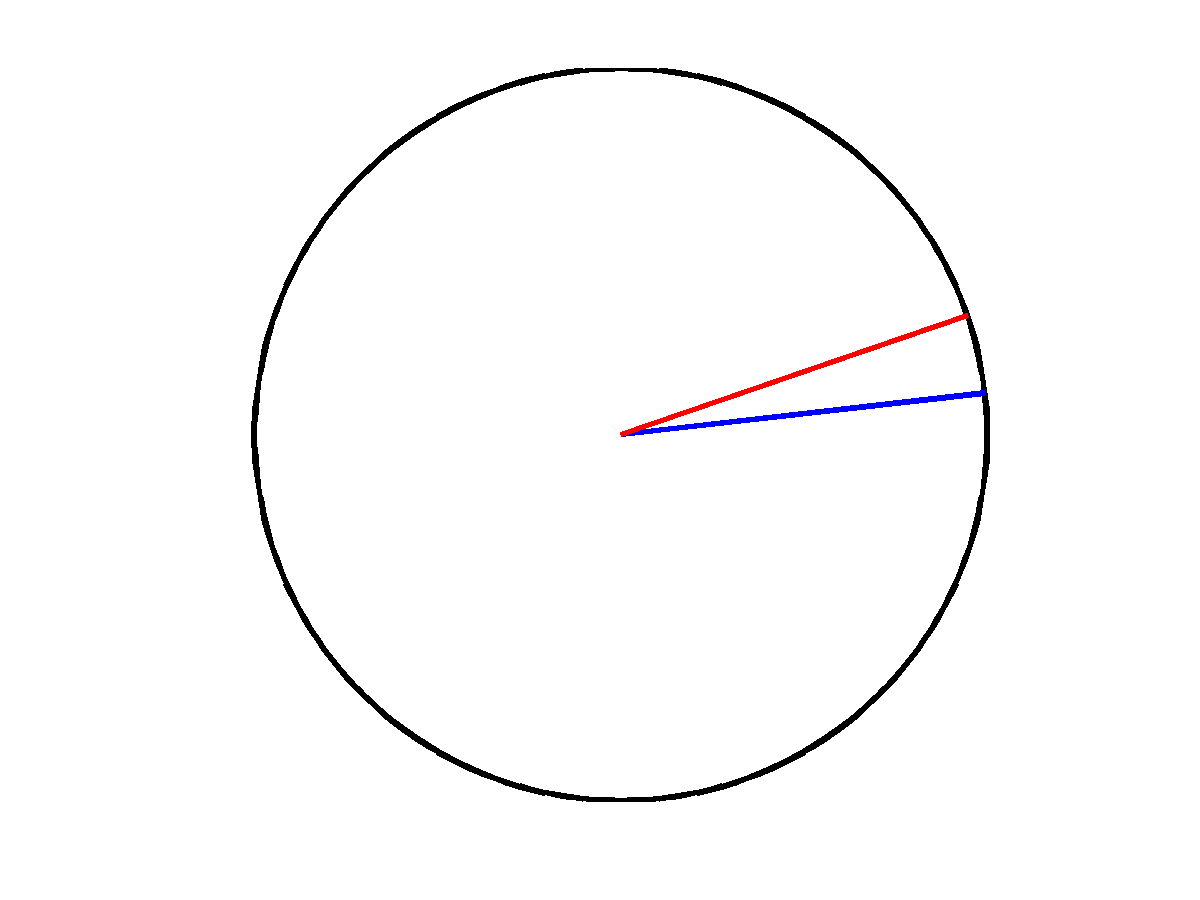} &
\includegraphics[width=0.51in]{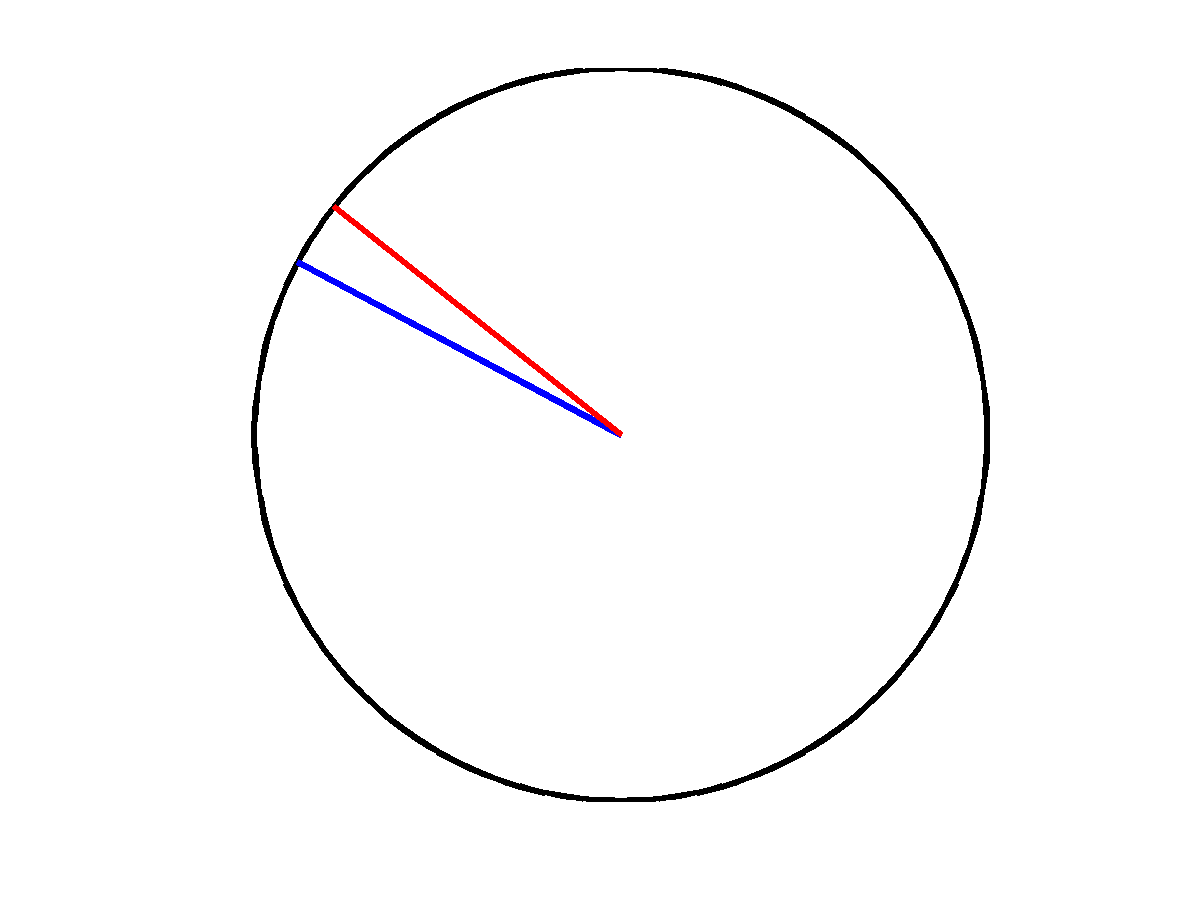} &
\includegraphics[width=0.51in]{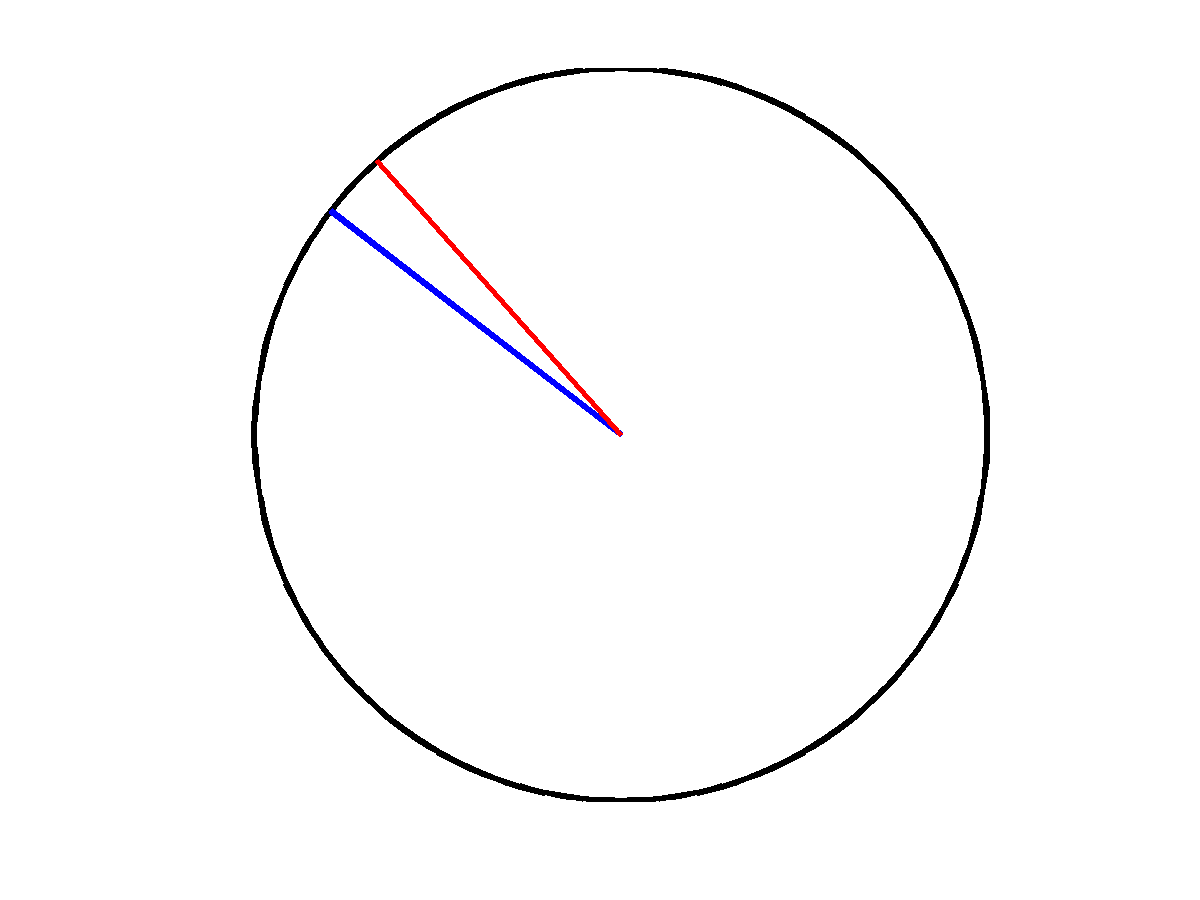} &
\includegraphics[width=0.51in]{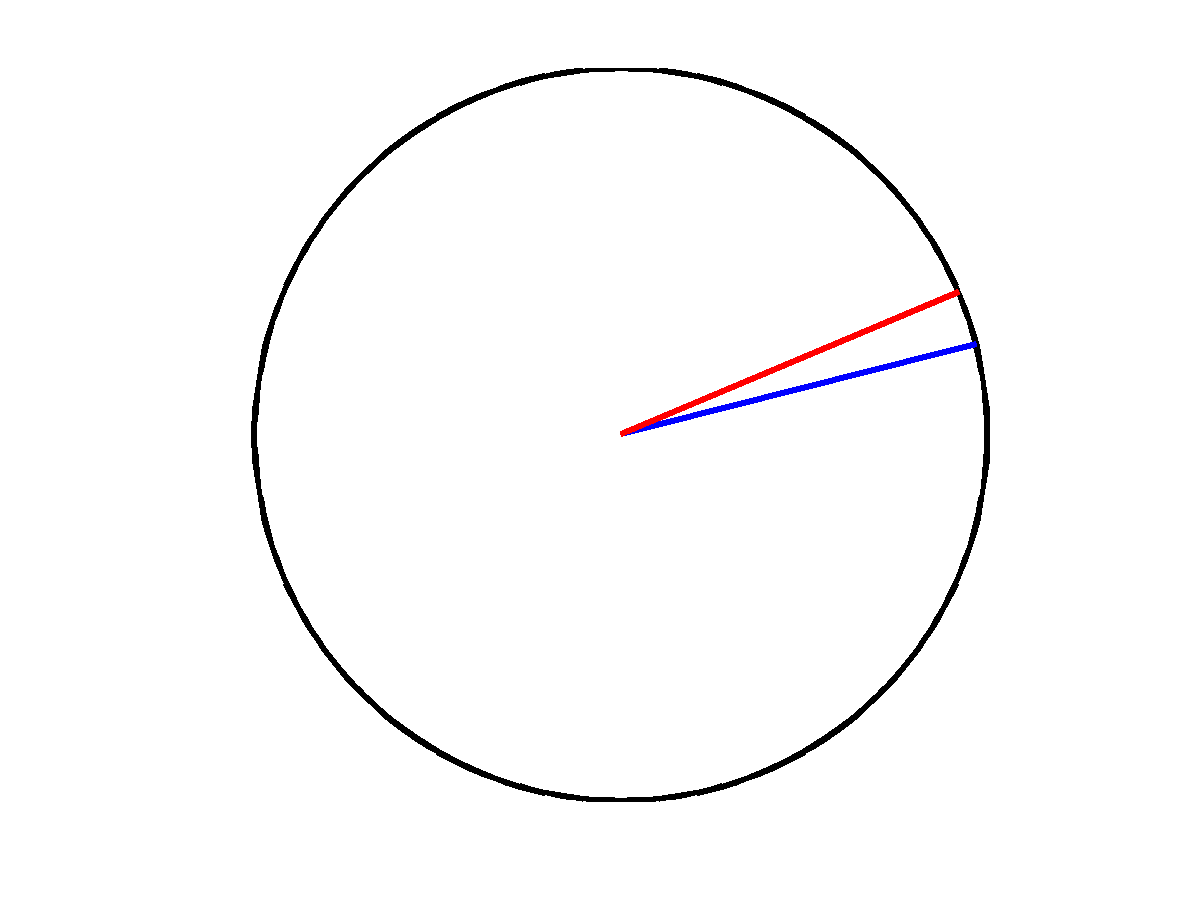} &
\includegraphics[width=0.51in]{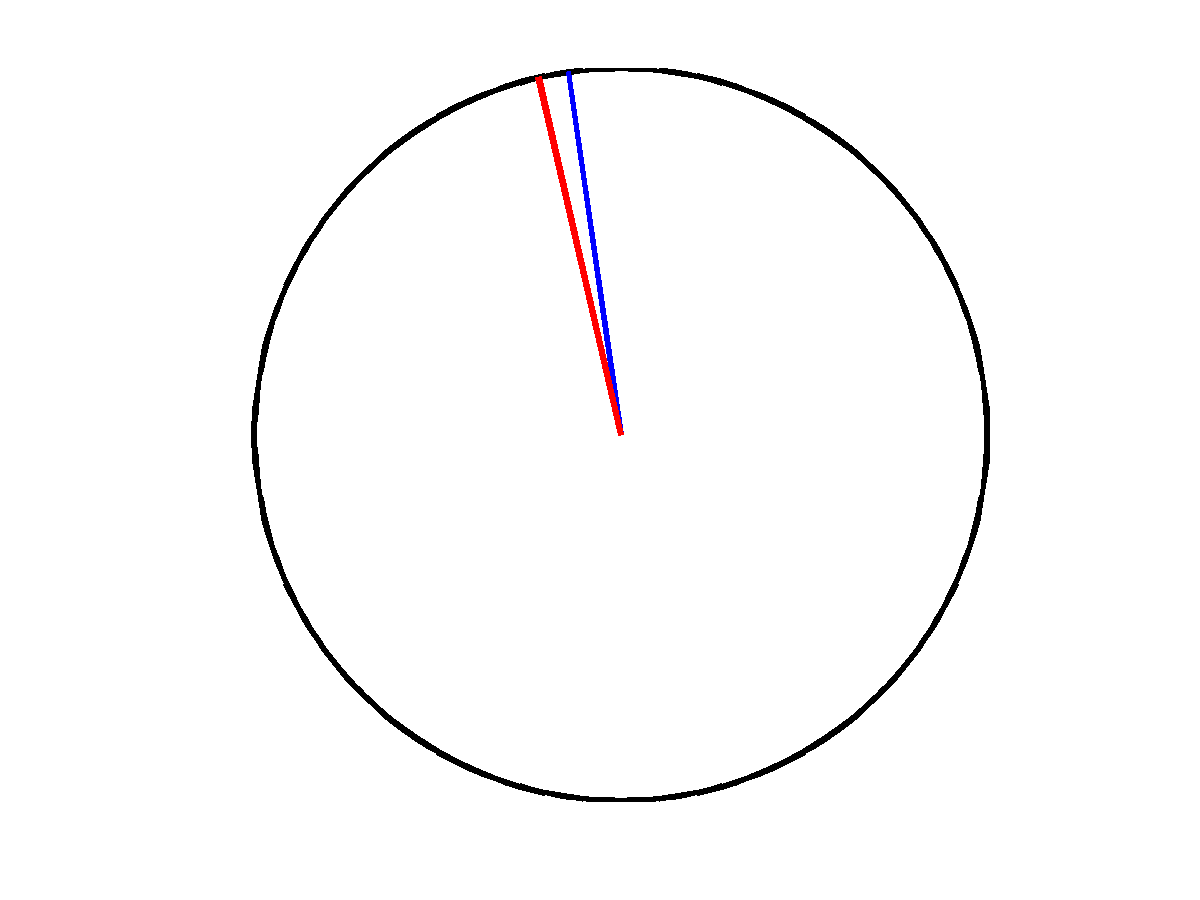} &
\includegraphics[width=0.51in]{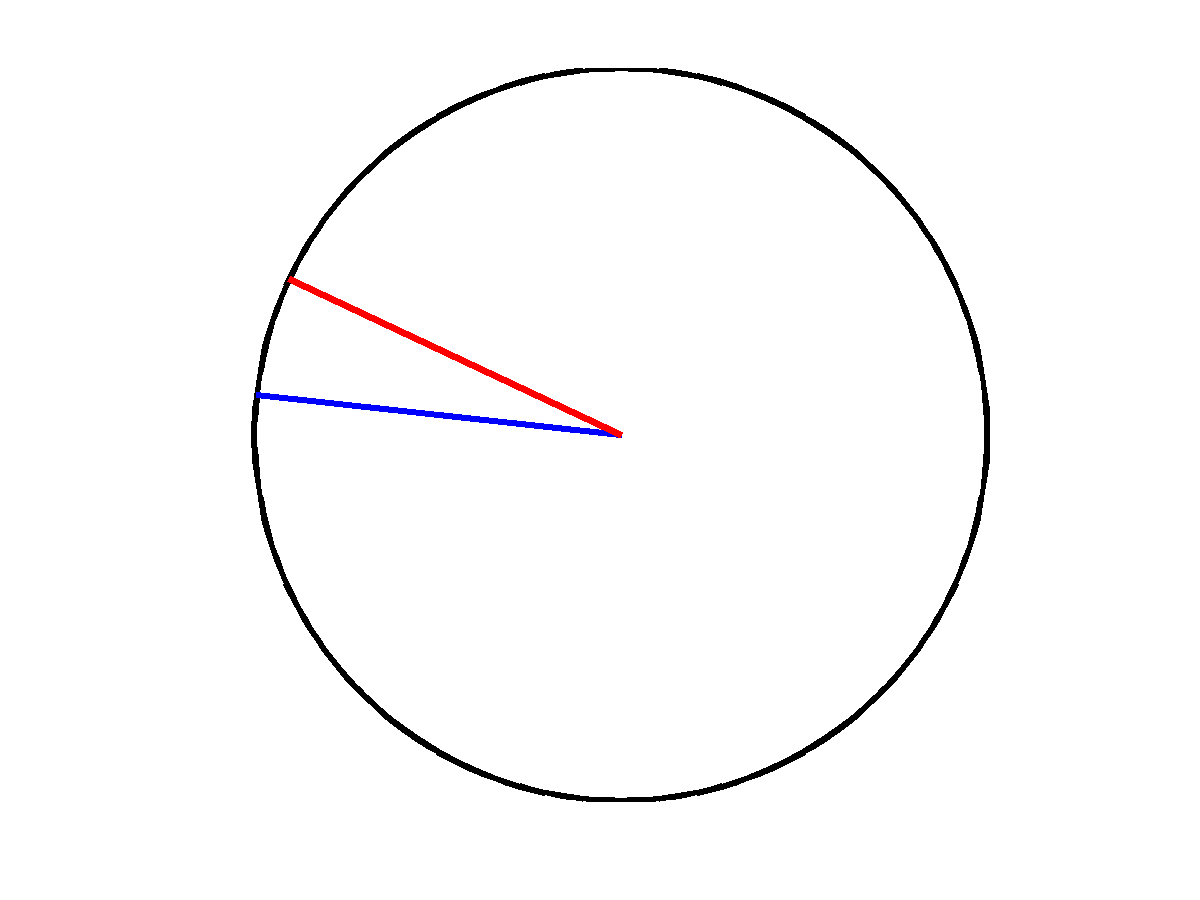} &
\includegraphics[width=0.51in]{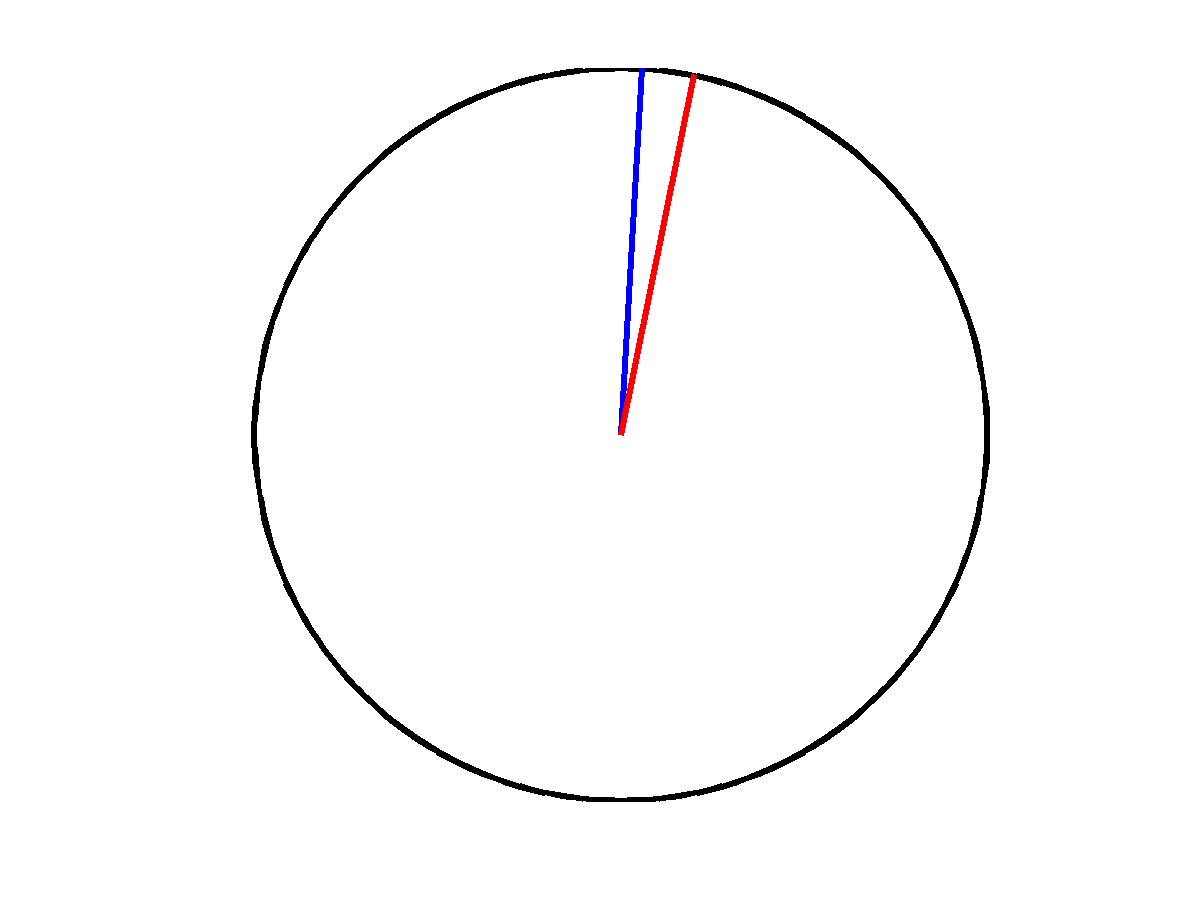} &
\includegraphics[width=0.51in]{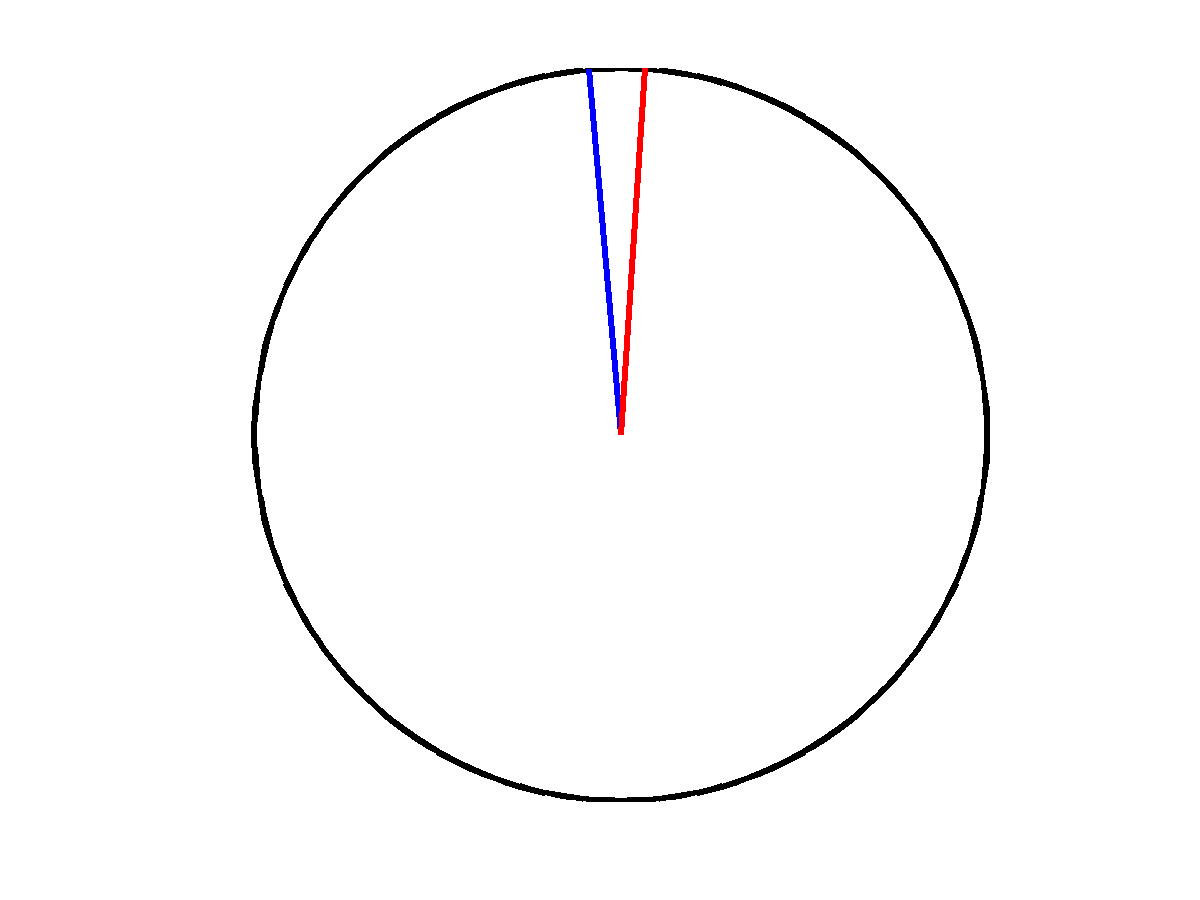}
\\
\includegraphics[width=0.51in]{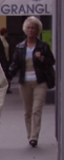} &
\includegraphics[width=0.51in]{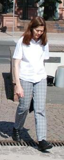} &
\includegraphics[width=0.51in]{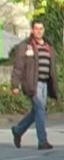} &
\includegraphics[width=0.51in]{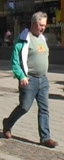} &
\includegraphics[width=0.51in]{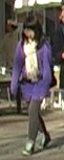} &
\includegraphics[width=0.51in]{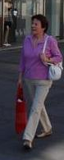} &
\includegraphics[width=0.51in]{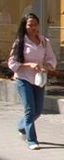} &
\includegraphics[width=0.51in]{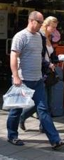} &
\includegraphics[width=0.51in]{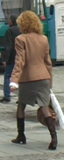} &
\includegraphics[width=0.51in]{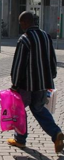}
\\
\includegraphics[width=0.51in]{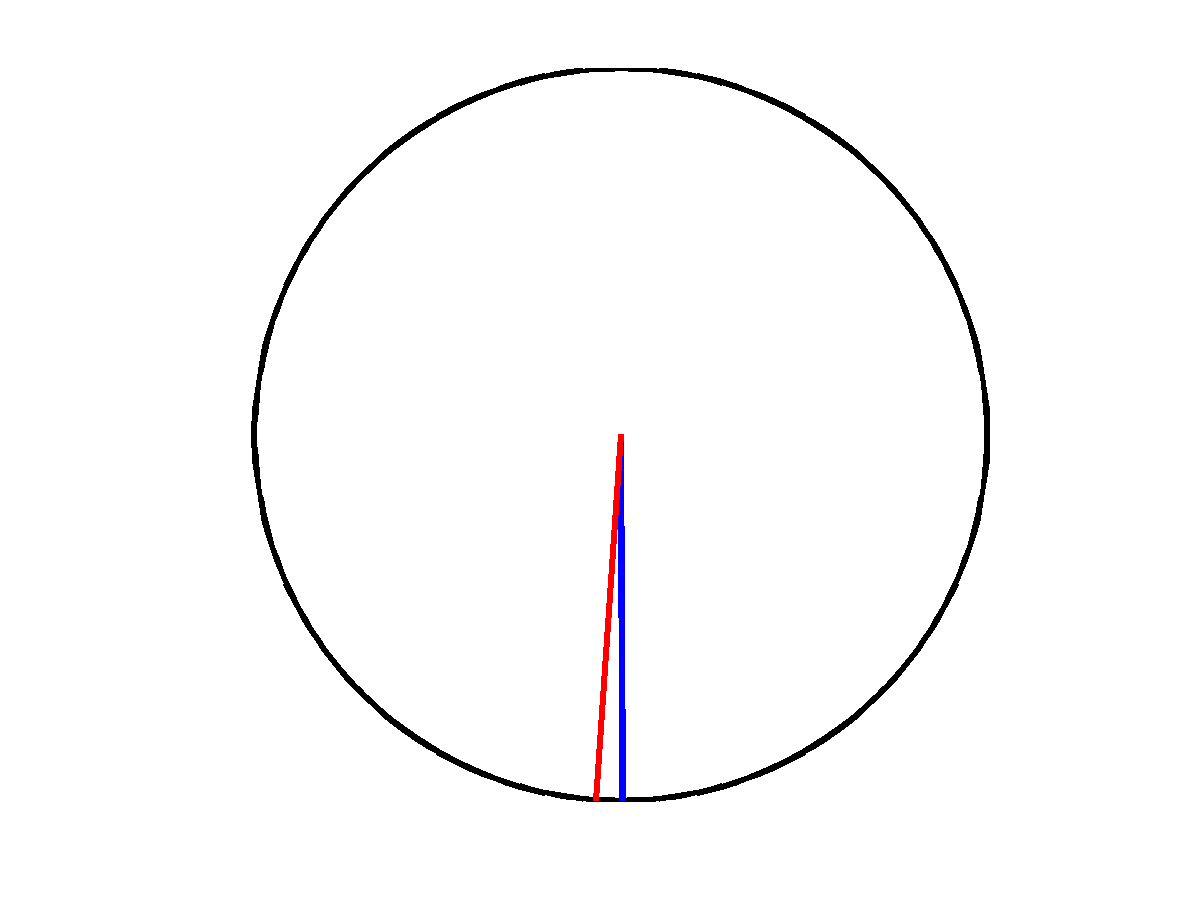} &
\includegraphics[width=0.51in]{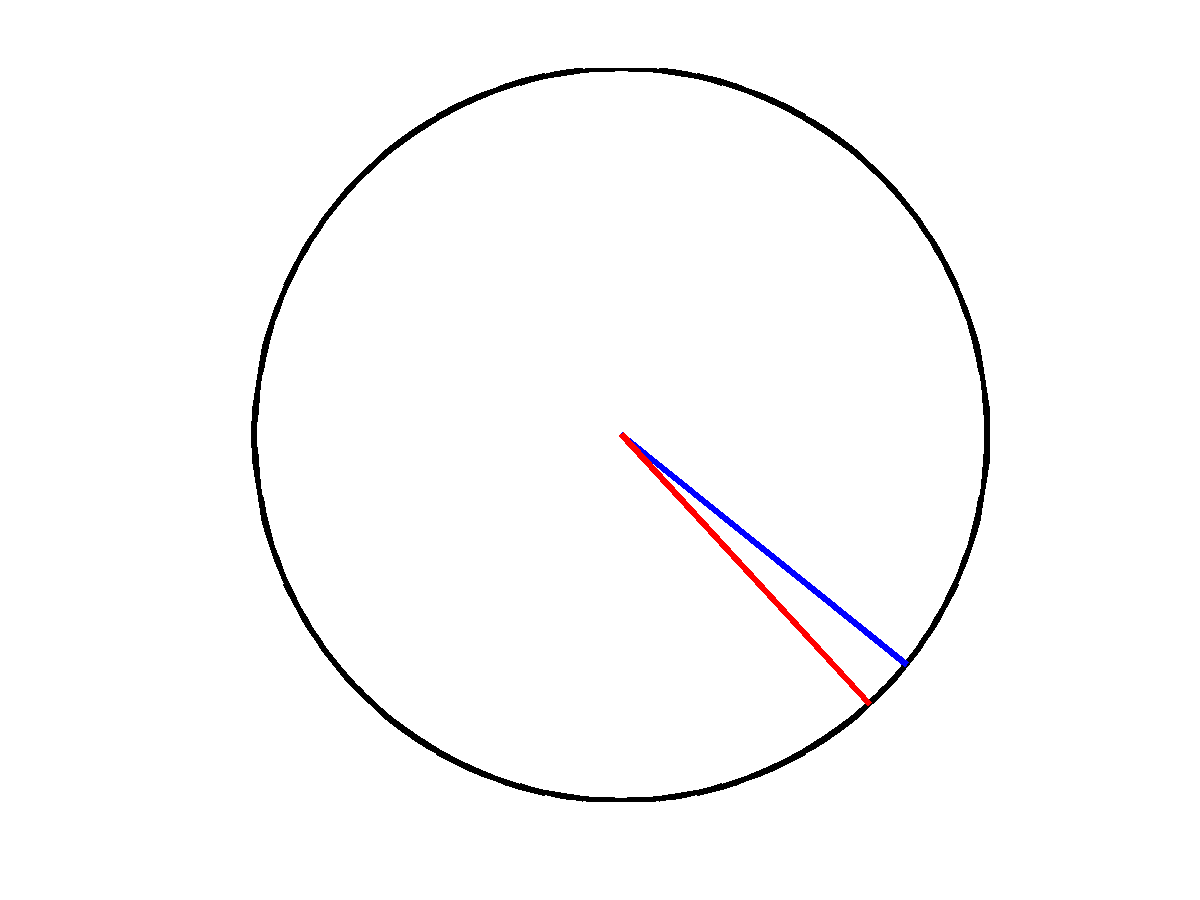} &
\includegraphics[width=0.51in]{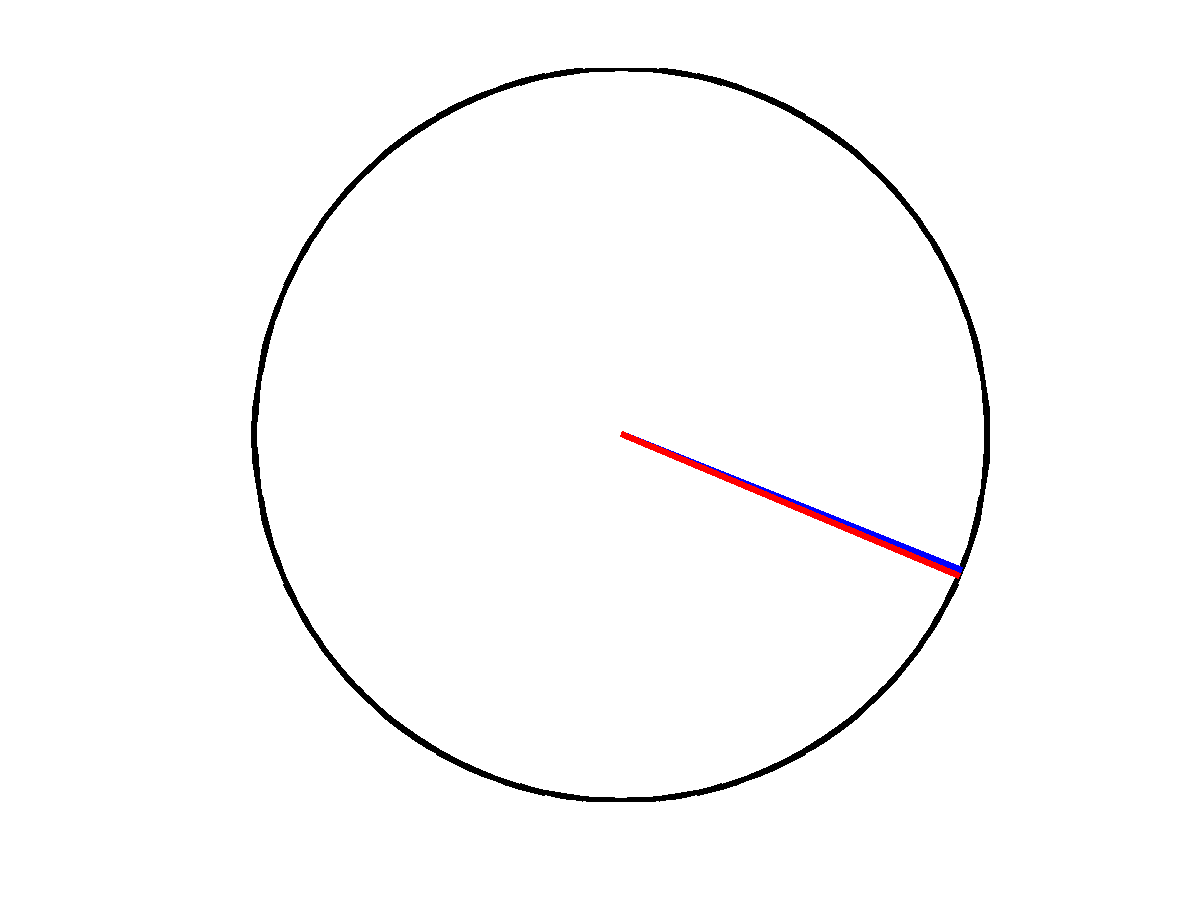} &
\includegraphics[width=0.51in]{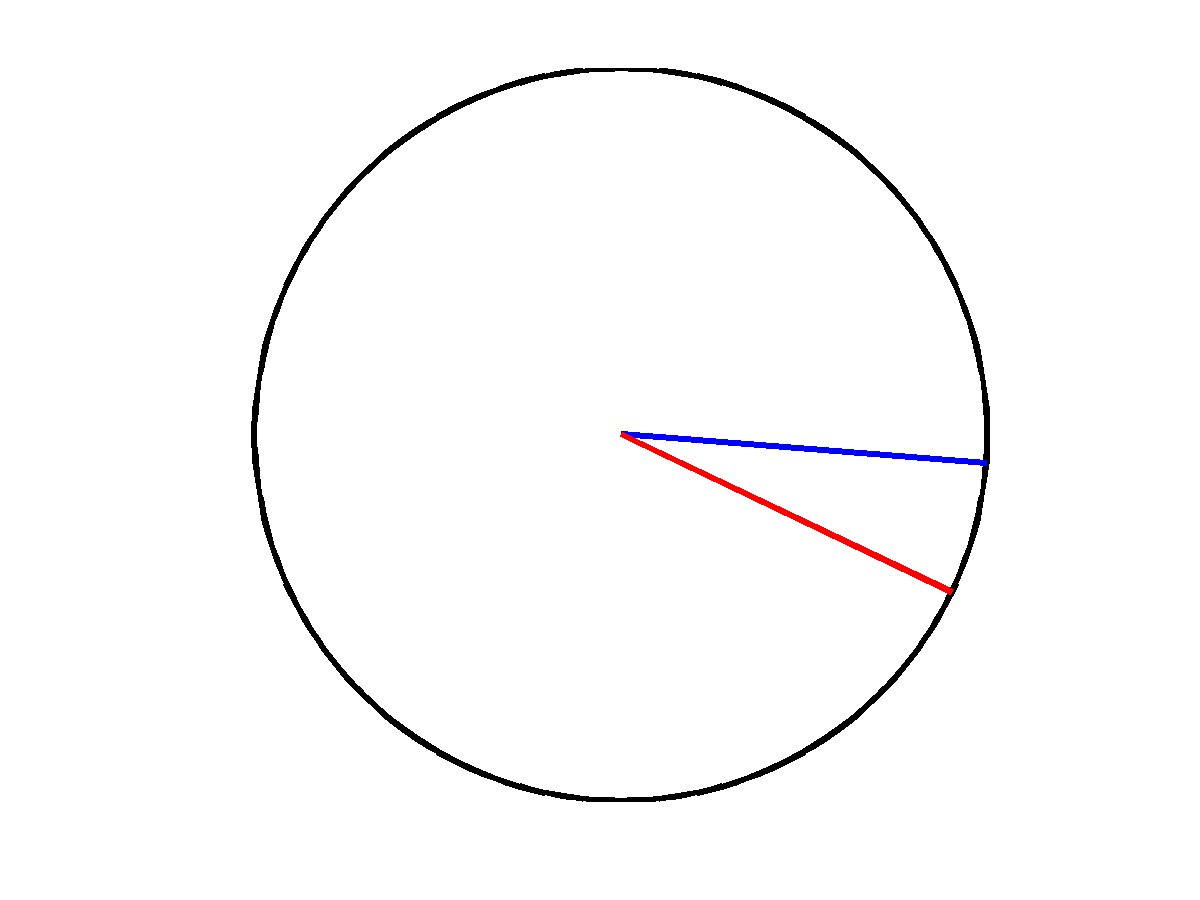} &
\includegraphics[width=0.51in]{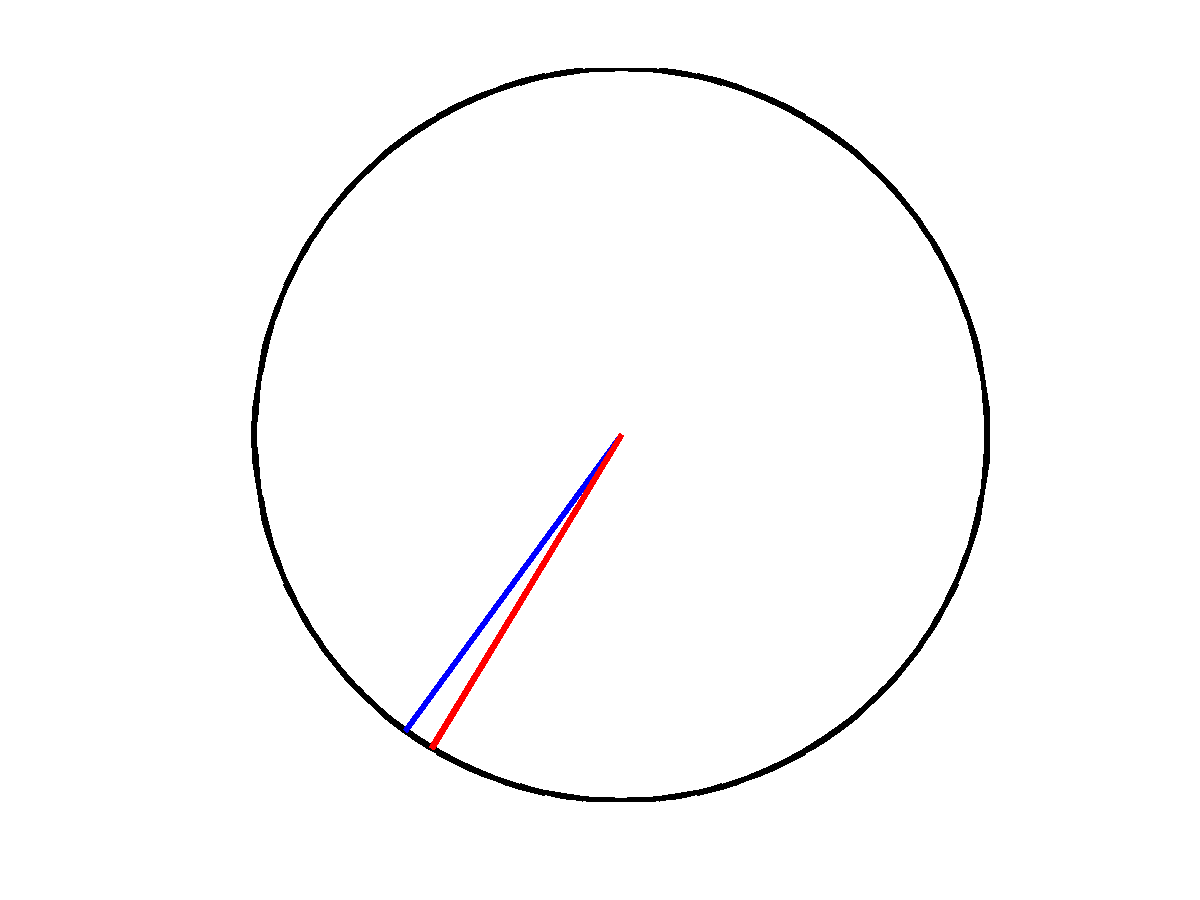} &
\includegraphics[width=0.51in]{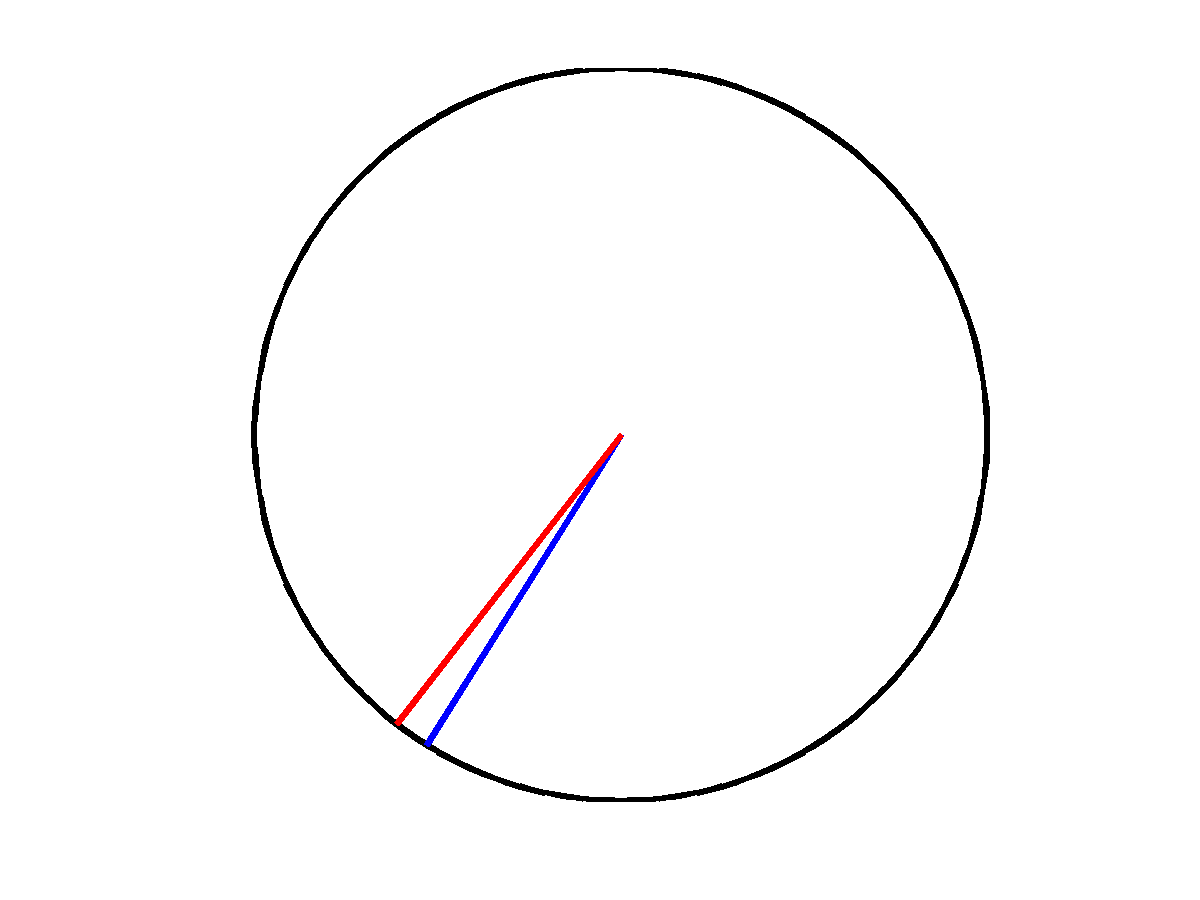} &
\includegraphics[width=0.51in]{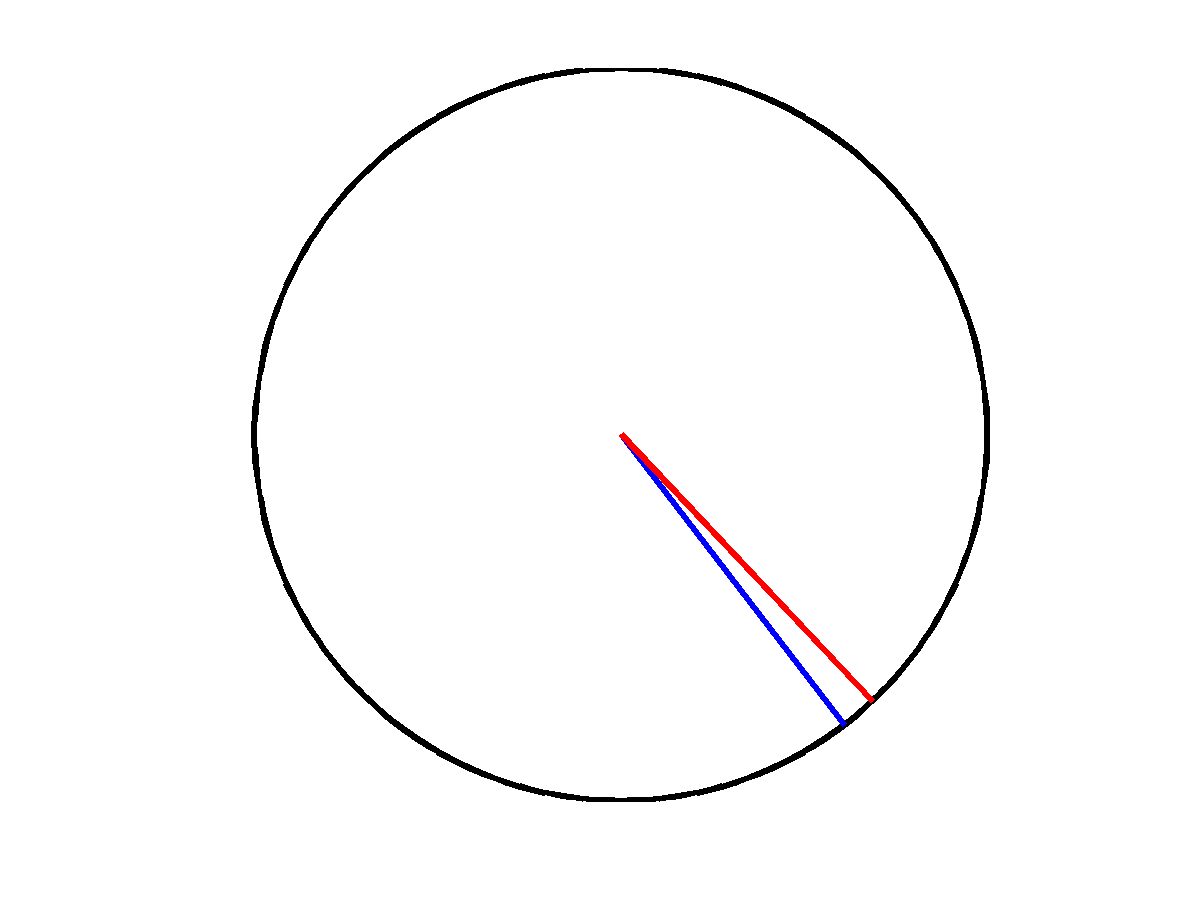} &
\includegraphics[width=0.51in]{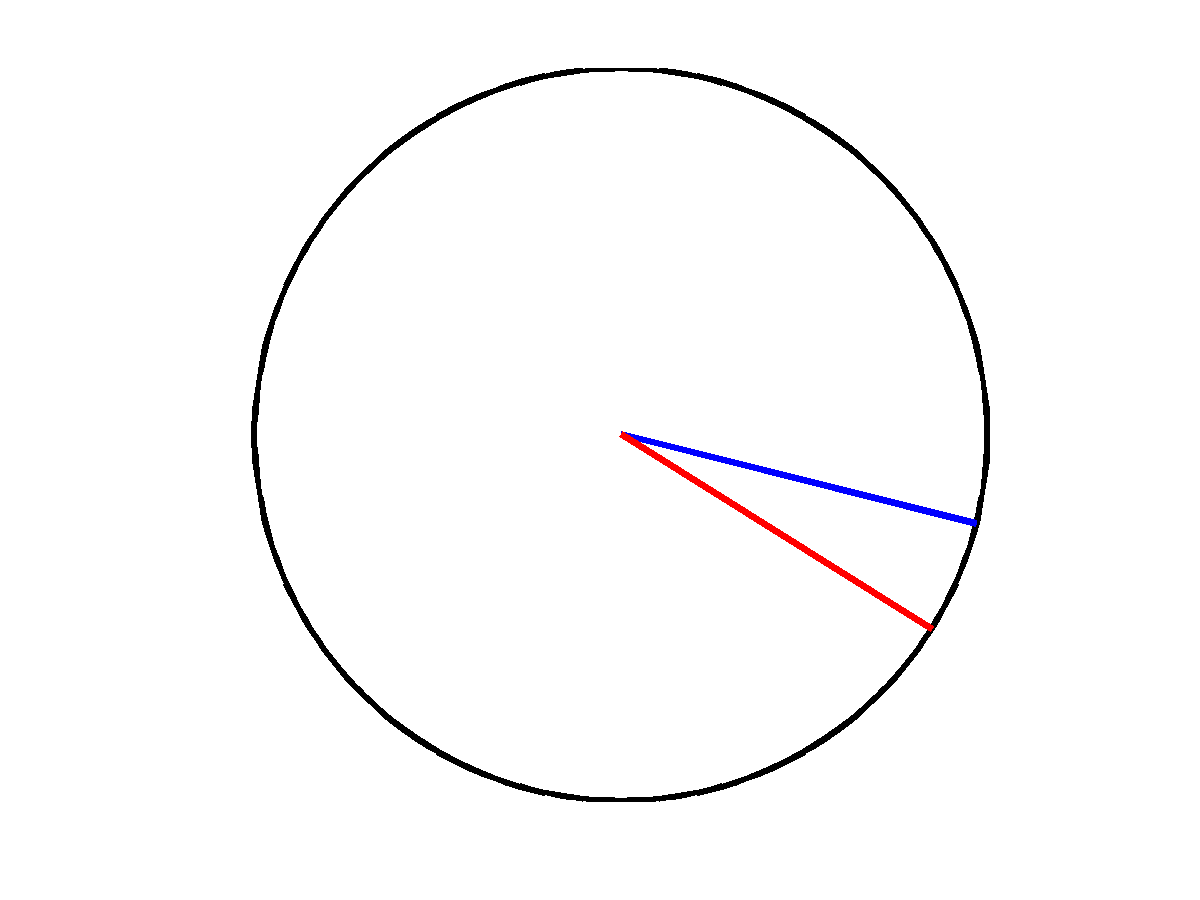} &
\includegraphics[width=0.51in]{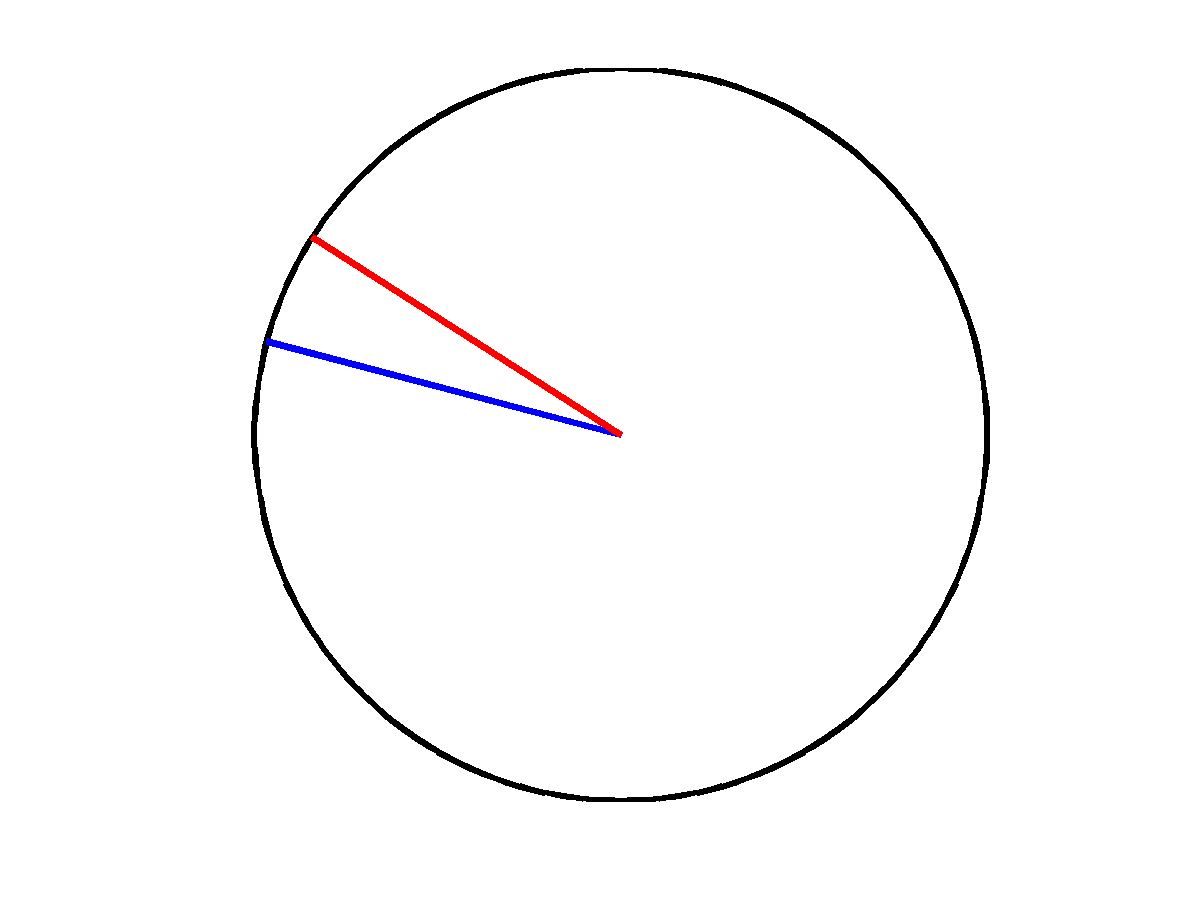} &
\includegraphics[width=0.51in]{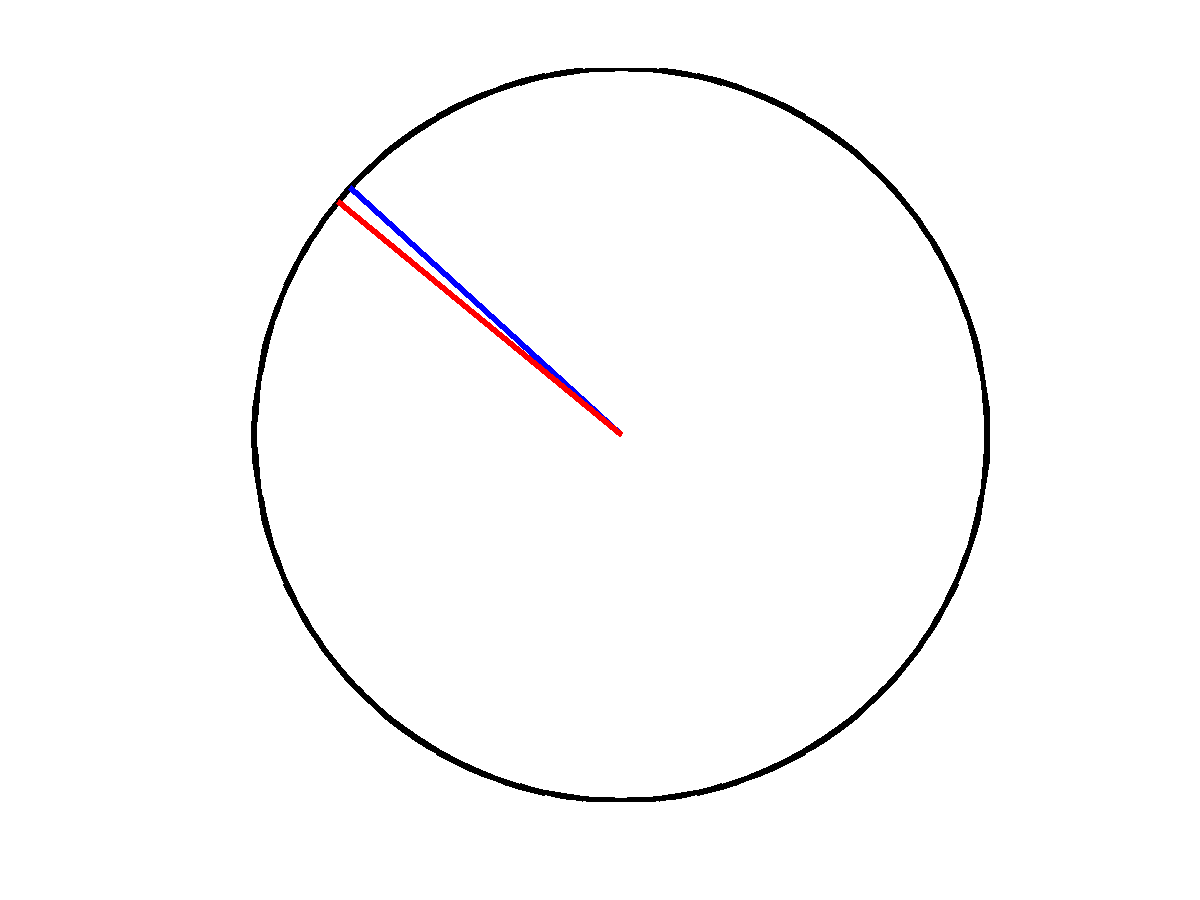}
\\
\includegraphics[width=0.51in]{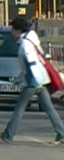} &
\includegraphics[width=0.51in]{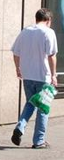} &
\includegraphics[width=0.51in]{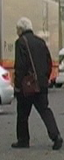} &
\includegraphics[width=0.51in]{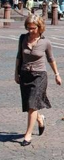} &
\includegraphics[width=0.51in]{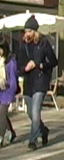} &
\includegraphics[width=0.51in]{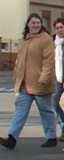} &
\includegraphics[width=0.51in]{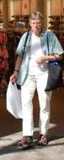} &
\includegraphics[width=0.51in]{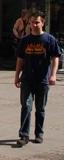} &
\includegraphics[width=0.51in]{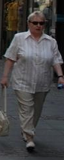} &
\includegraphics[width=0.51in]{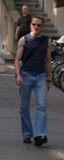}
\\
\includegraphics[width=0.51in]{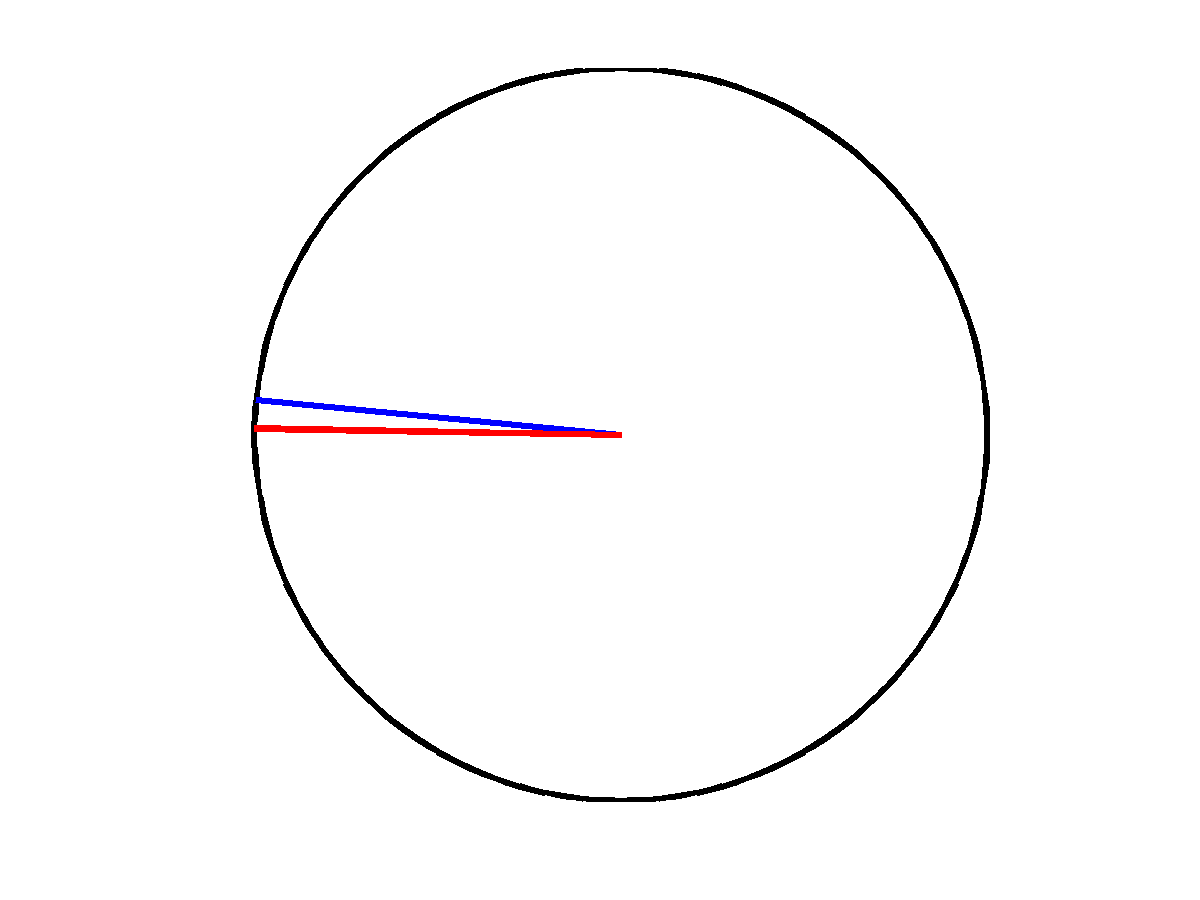} &
\includegraphics[width=0.51in]{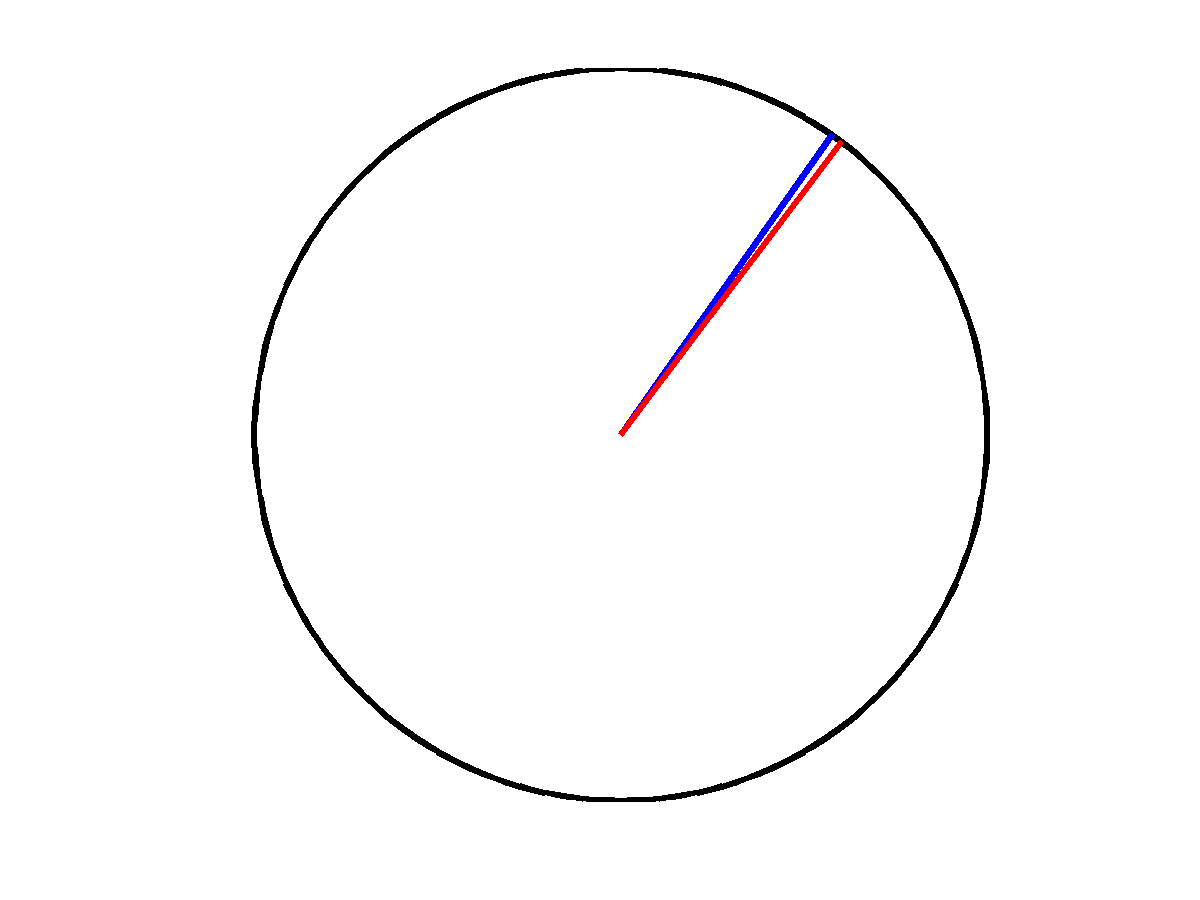} &
\includegraphics[width=0.51in]{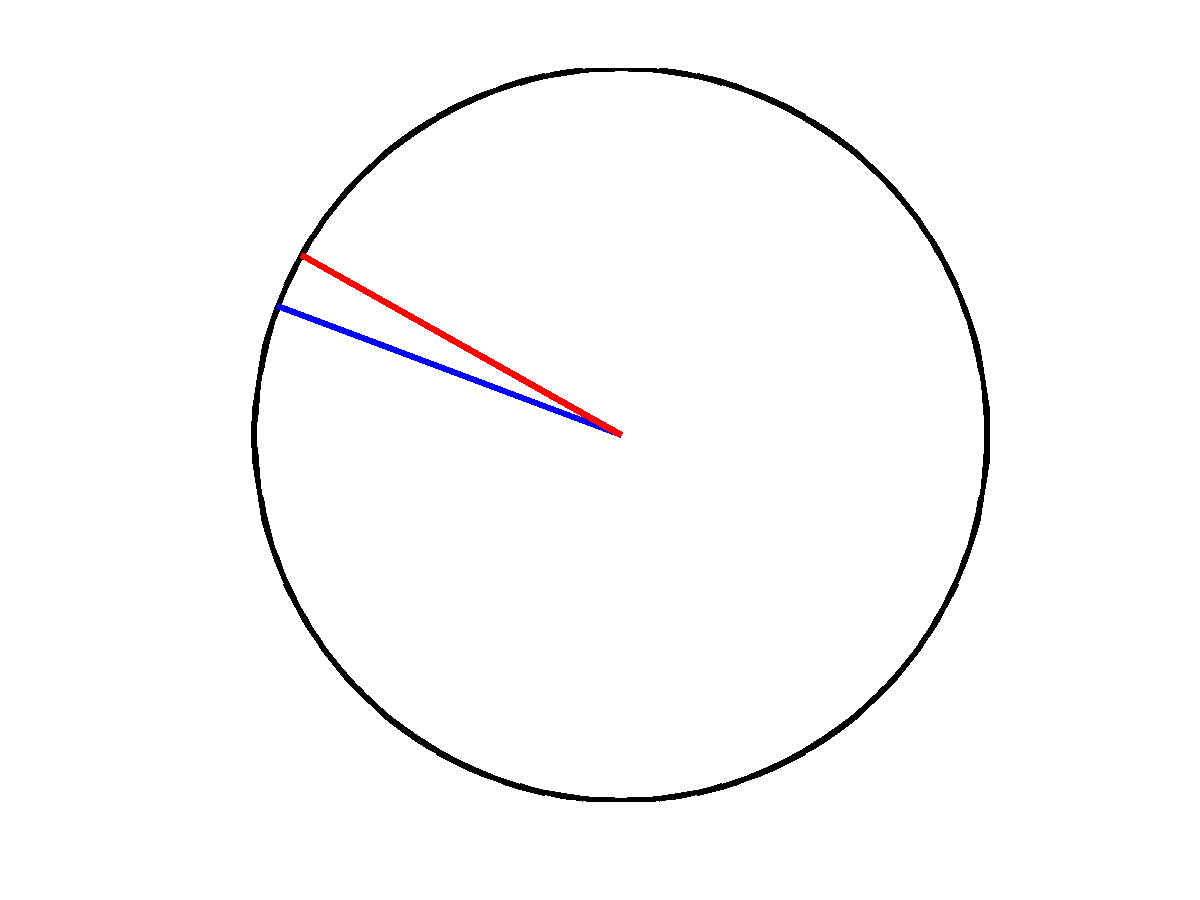} &
\includegraphics[width=0.51in]{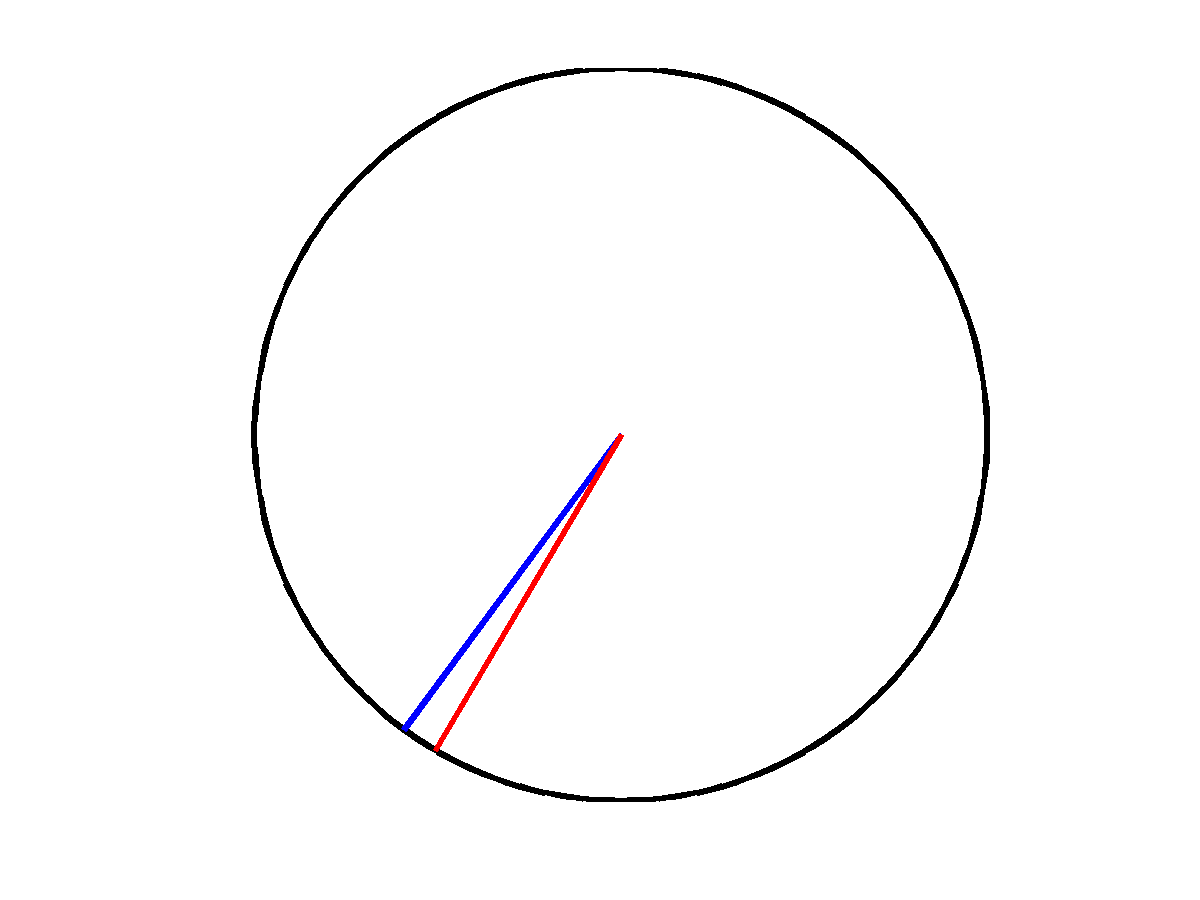} &
\includegraphics[width=0.51in]{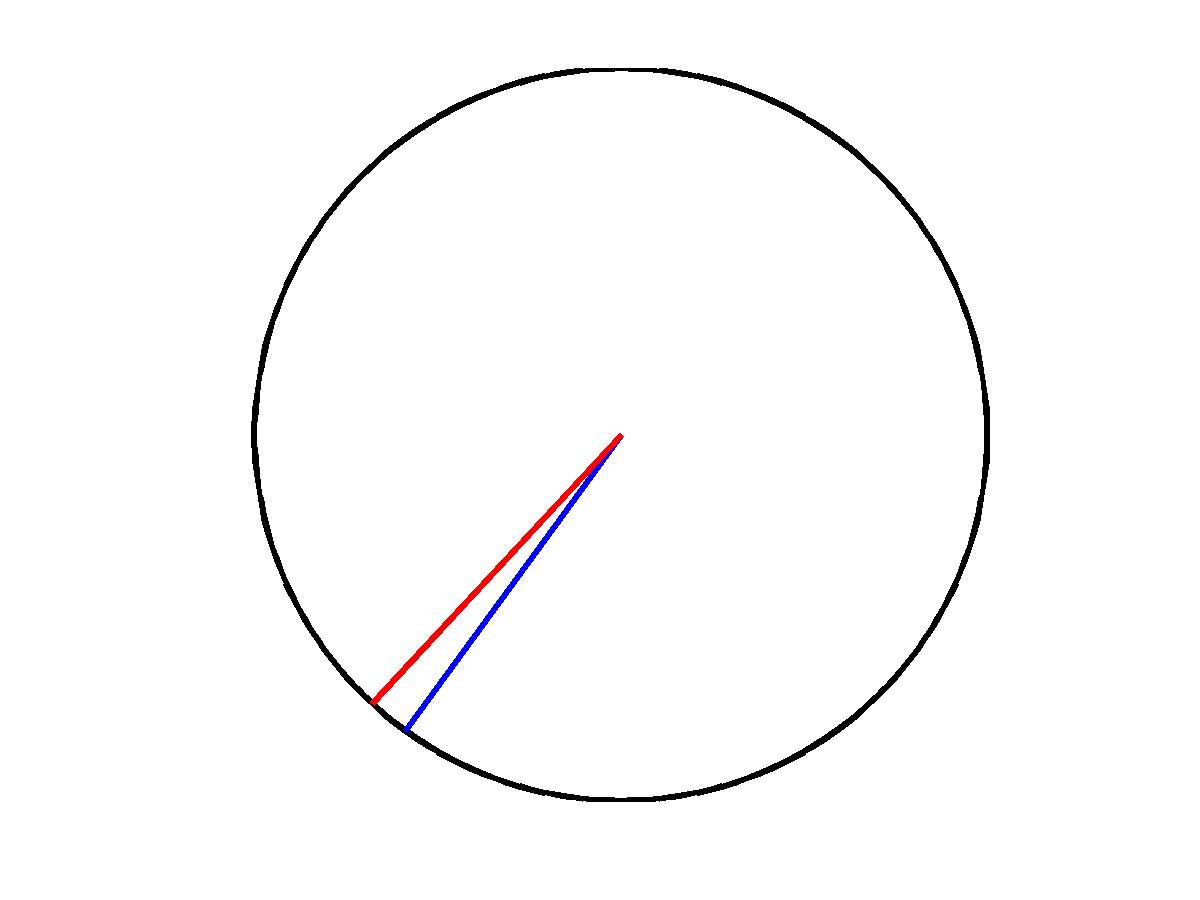} &
\includegraphics[width=0.51in]{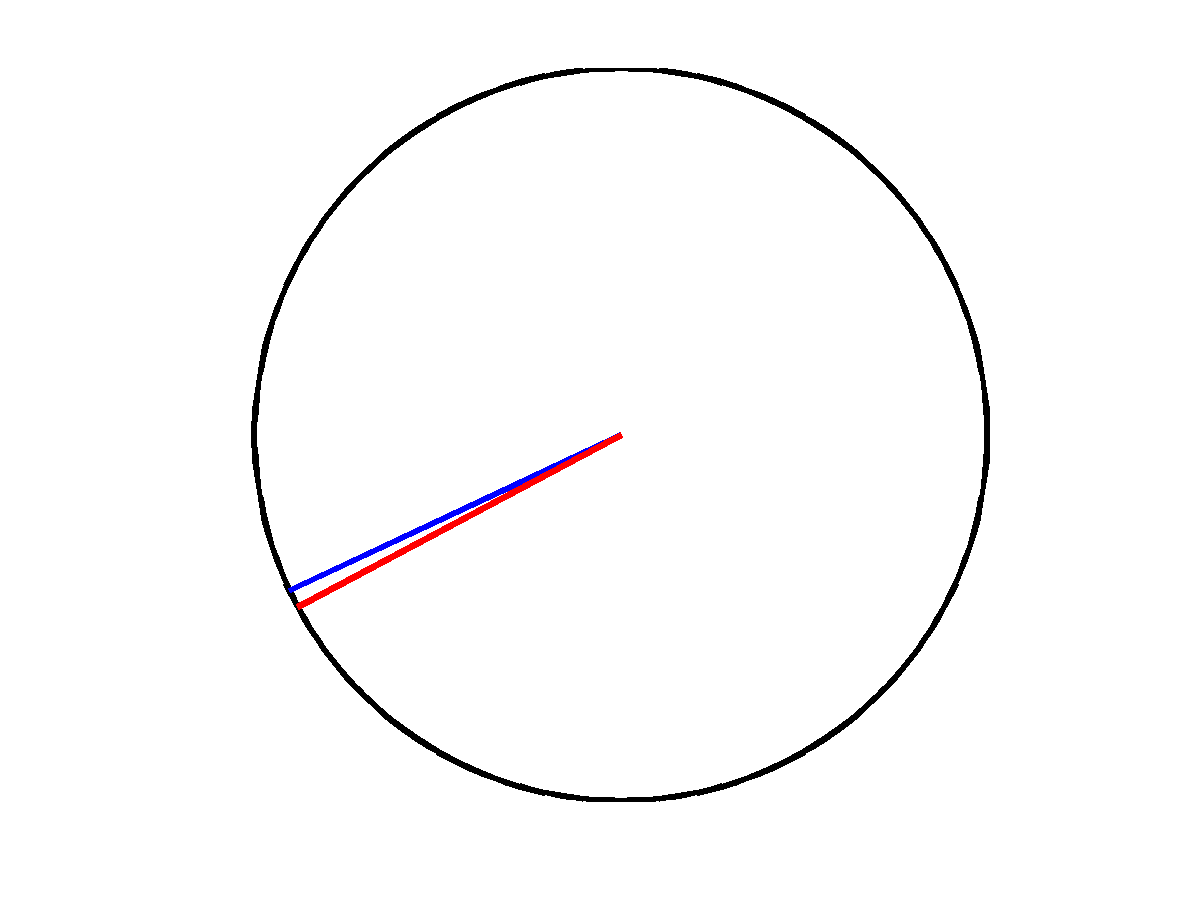} &
\includegraphics[width=0.51in]{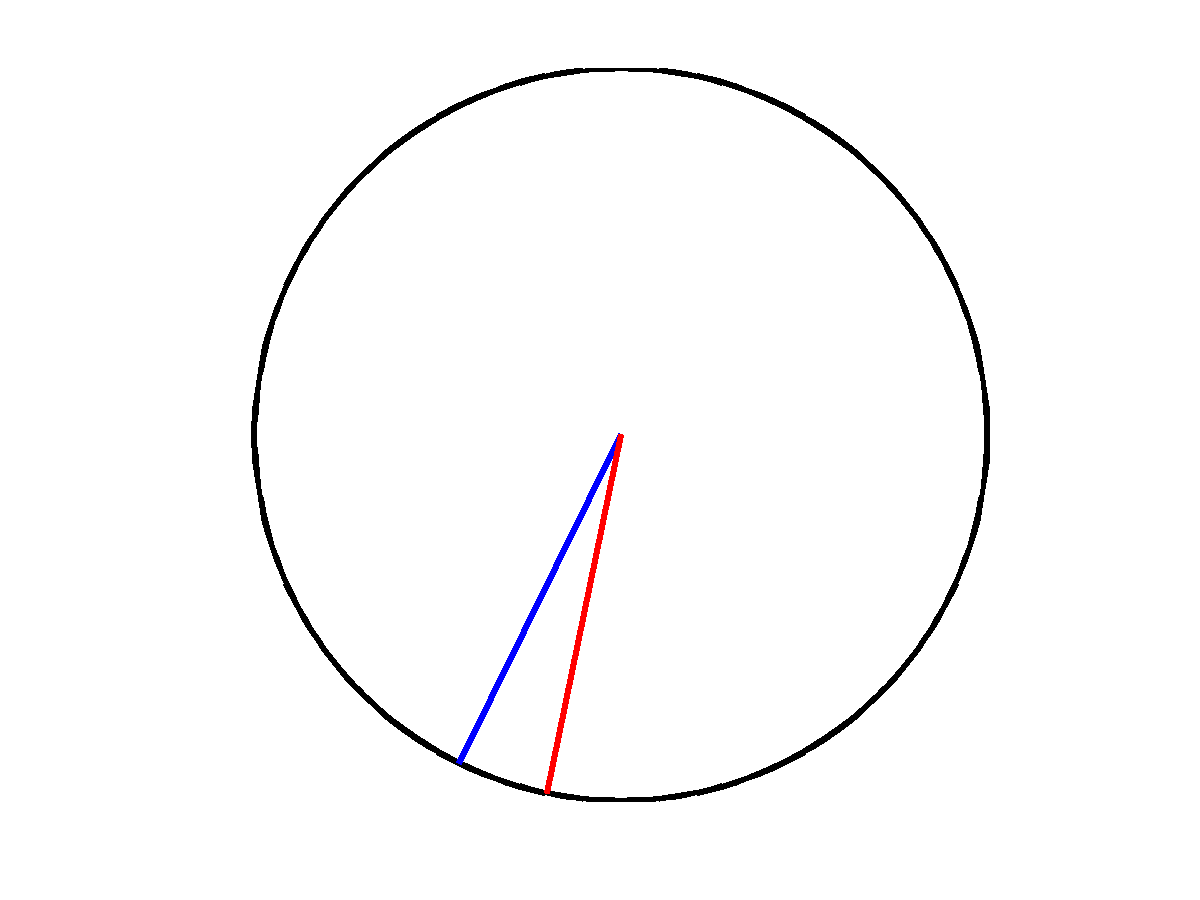} &
\includegraphics[width=0.51in]{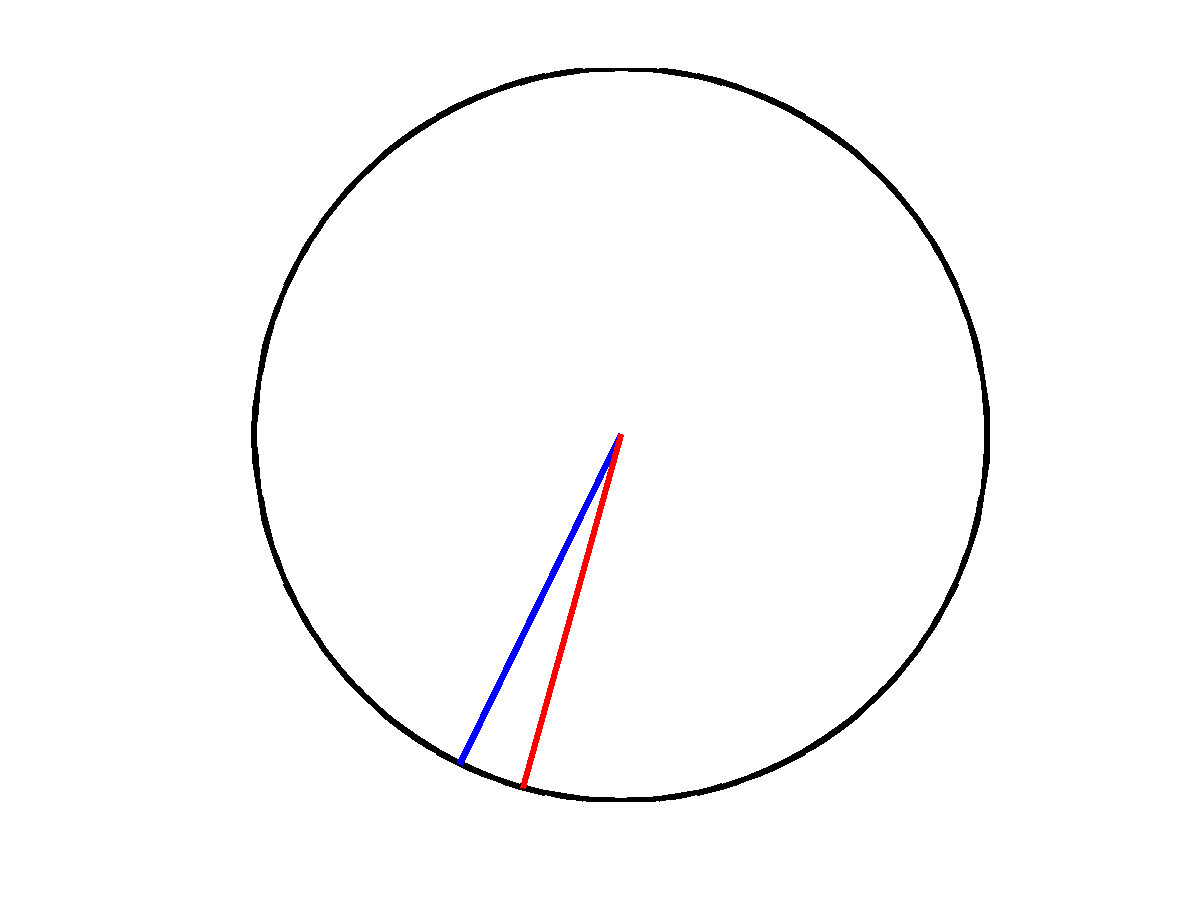} &
\includegraphics[width=0.51in]{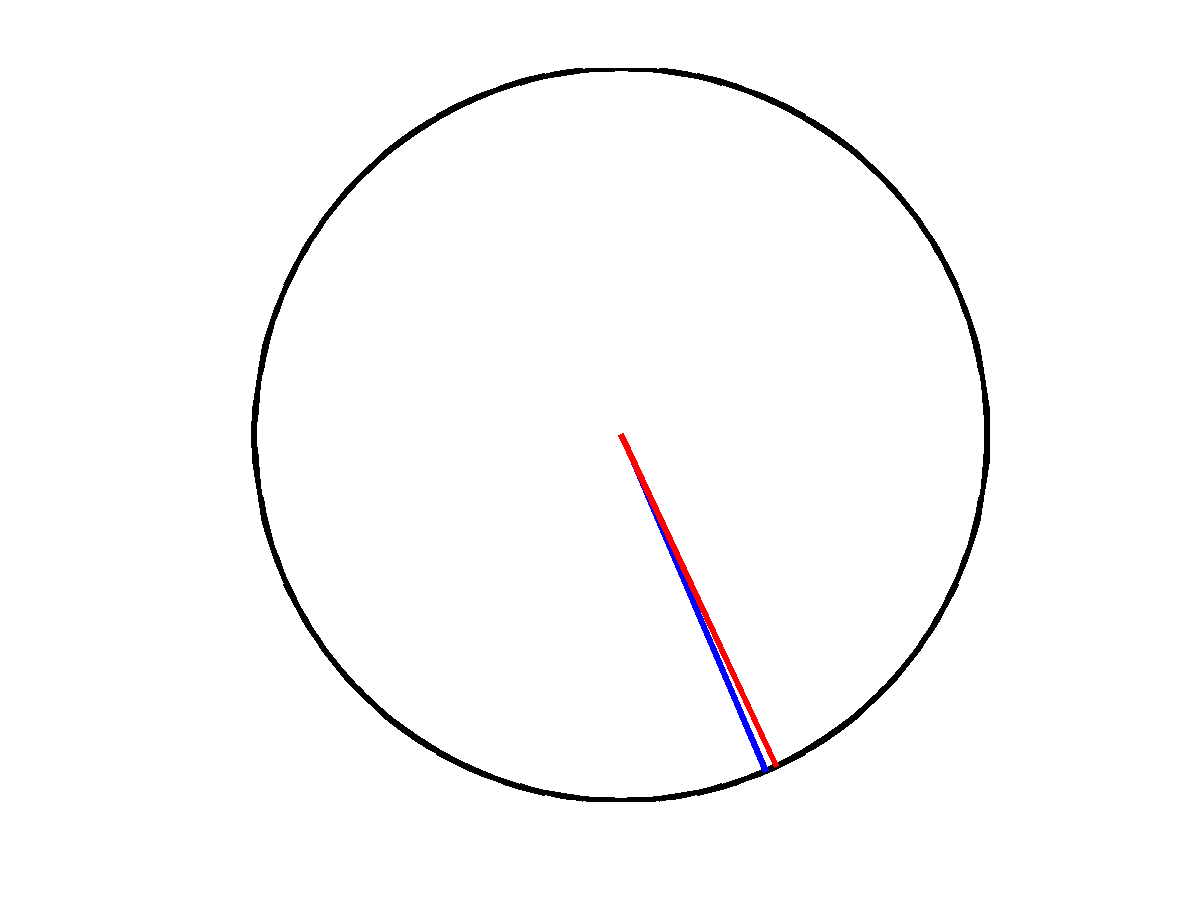} &
\includegraphics[width=0.51in]{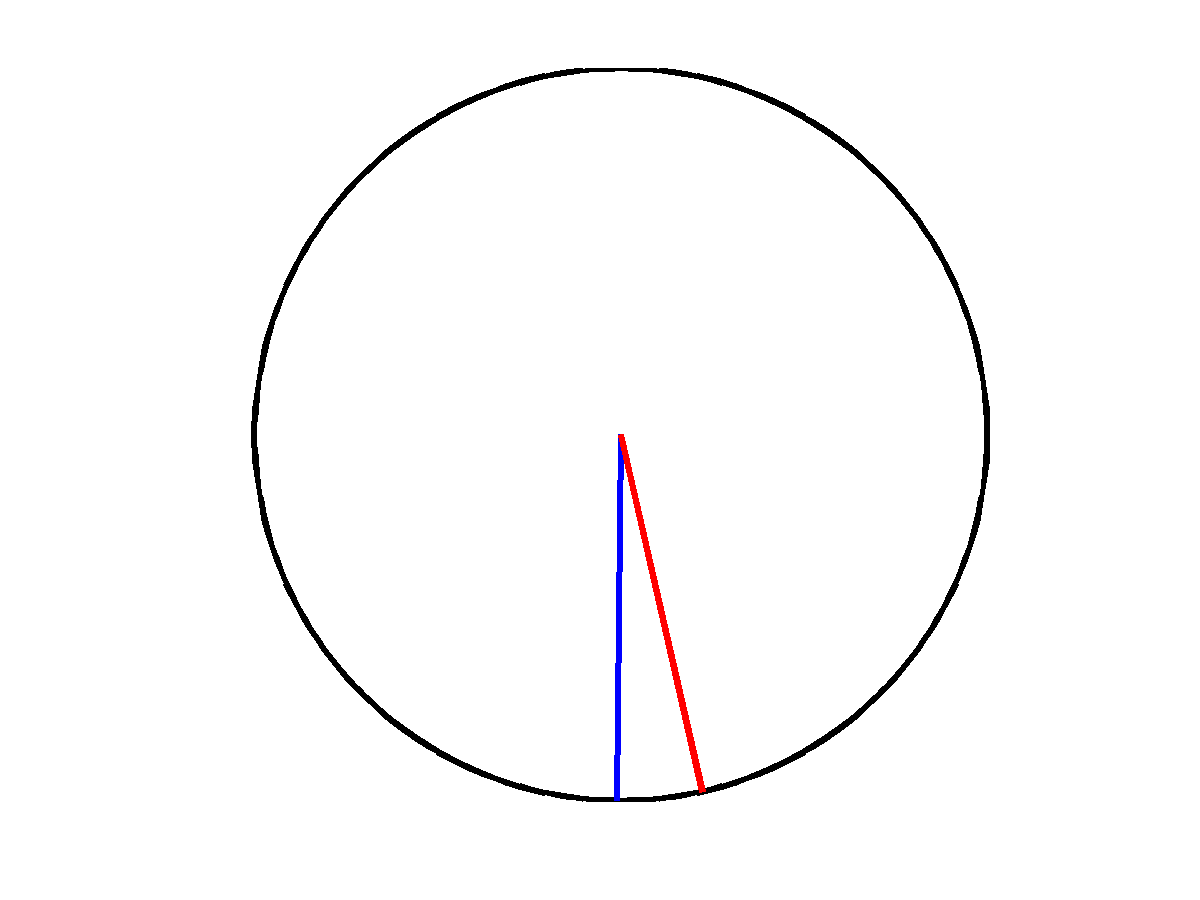}
\\
\includegraphics[width=0.51in]{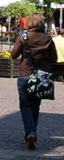} &
\includegraphics[width=0.51in]{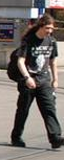} &
\includegraphics[width=0.51in]{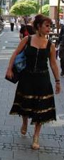} &
\includegraphics[width=0.51in]{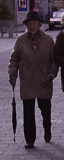} &
\includegraphics[width=0.51in]{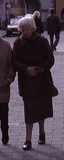} &
\includegraphics[width=0.51in]{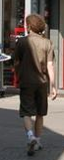} &
\includegraphics[width=0.51in]{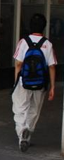} &
\includegraphics[width=0.51in]{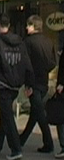} &
\includegraphics[width=0.51in]{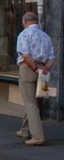} &
\includegraphics[width=0.51in]{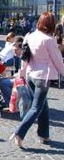}
\\
\includegraphics[width=0.51in]{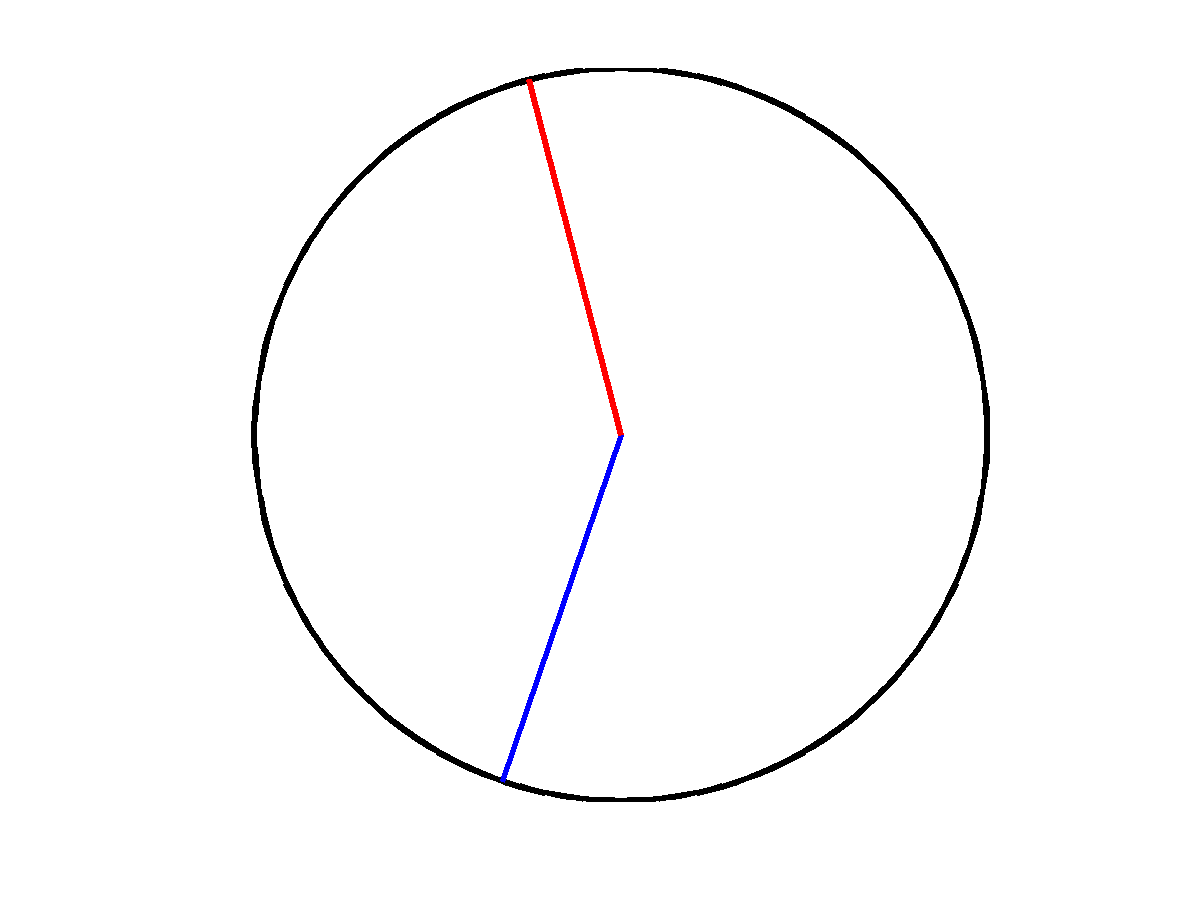} &
\includegraphics[width=0.51in]{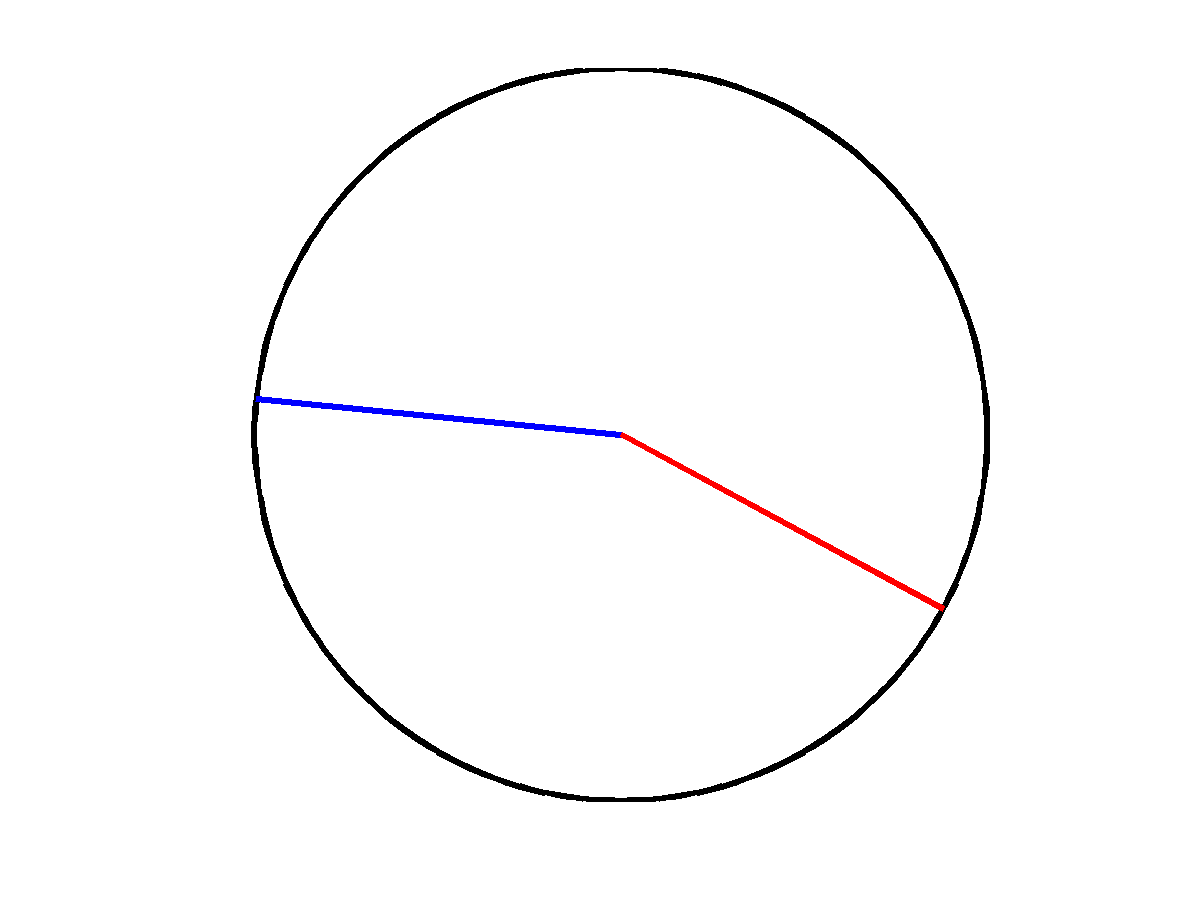} &
\includegraphics[width=0.51in]{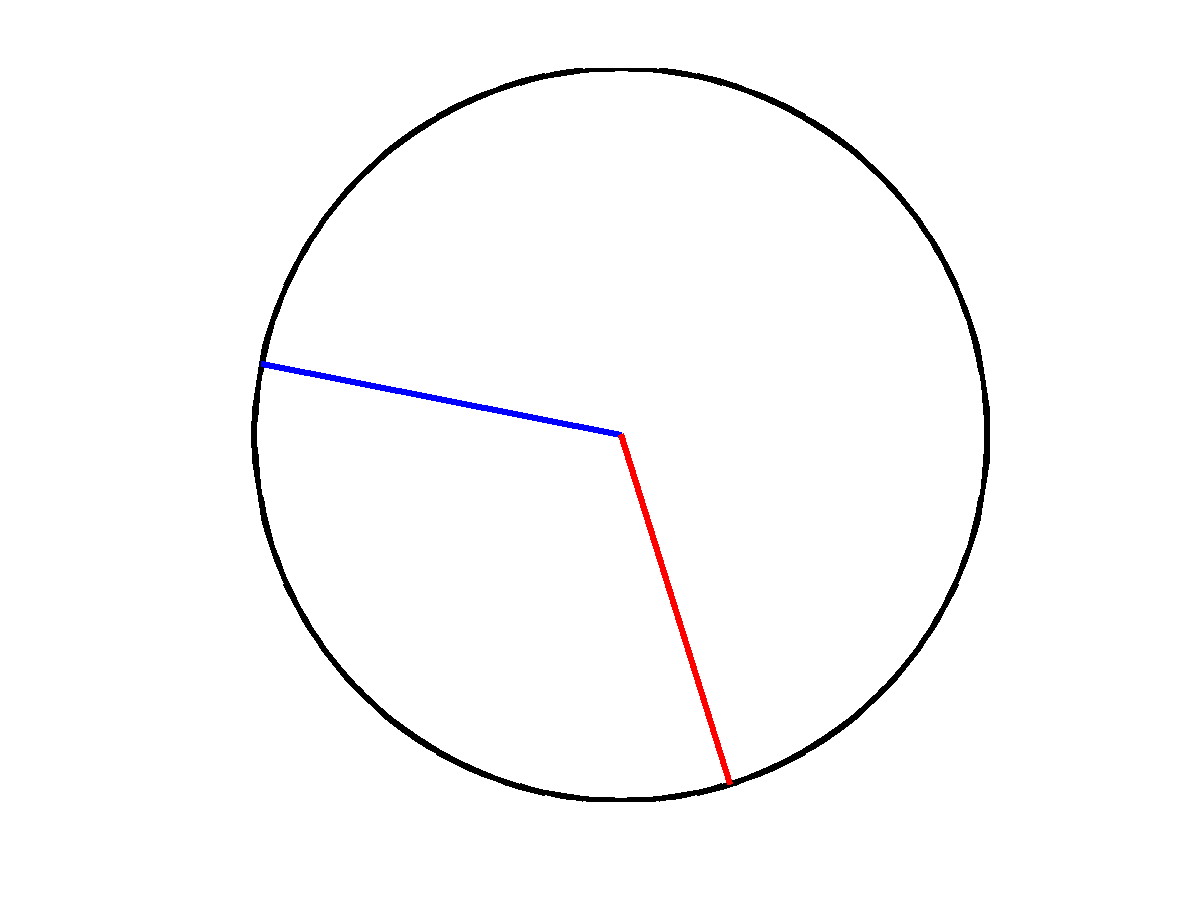} &
\includegraphics[width=0.51in]{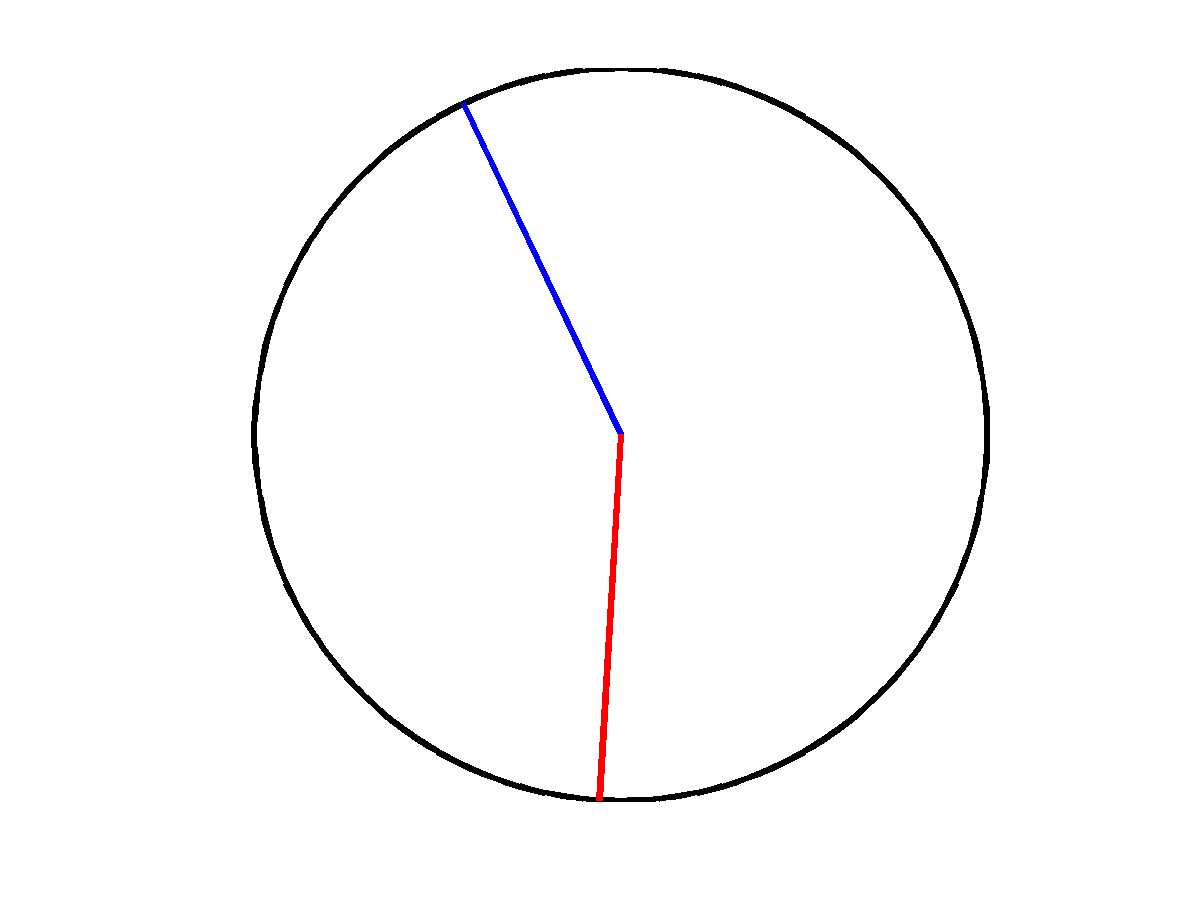} &
\includegraphics[width=0.51in]{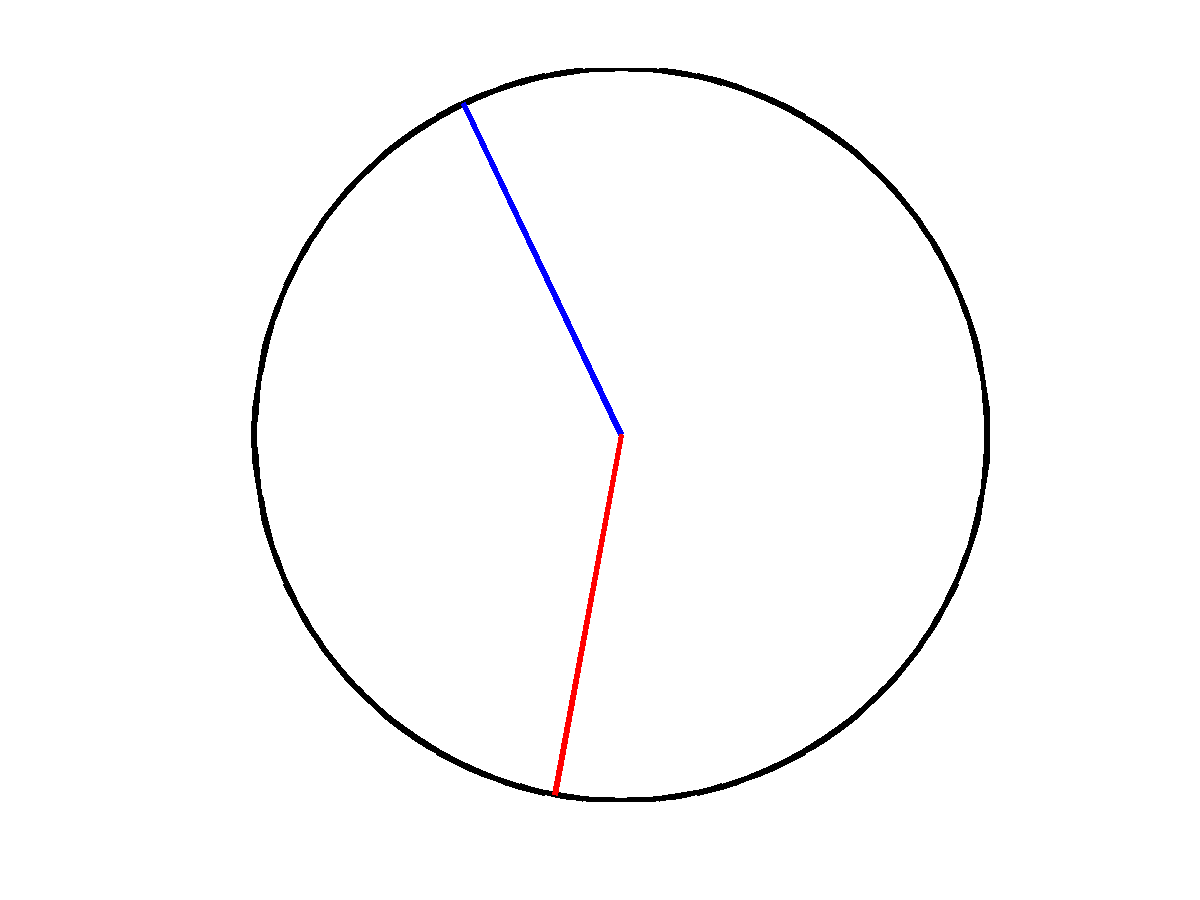} &
\includegraphics[width=0.51in]{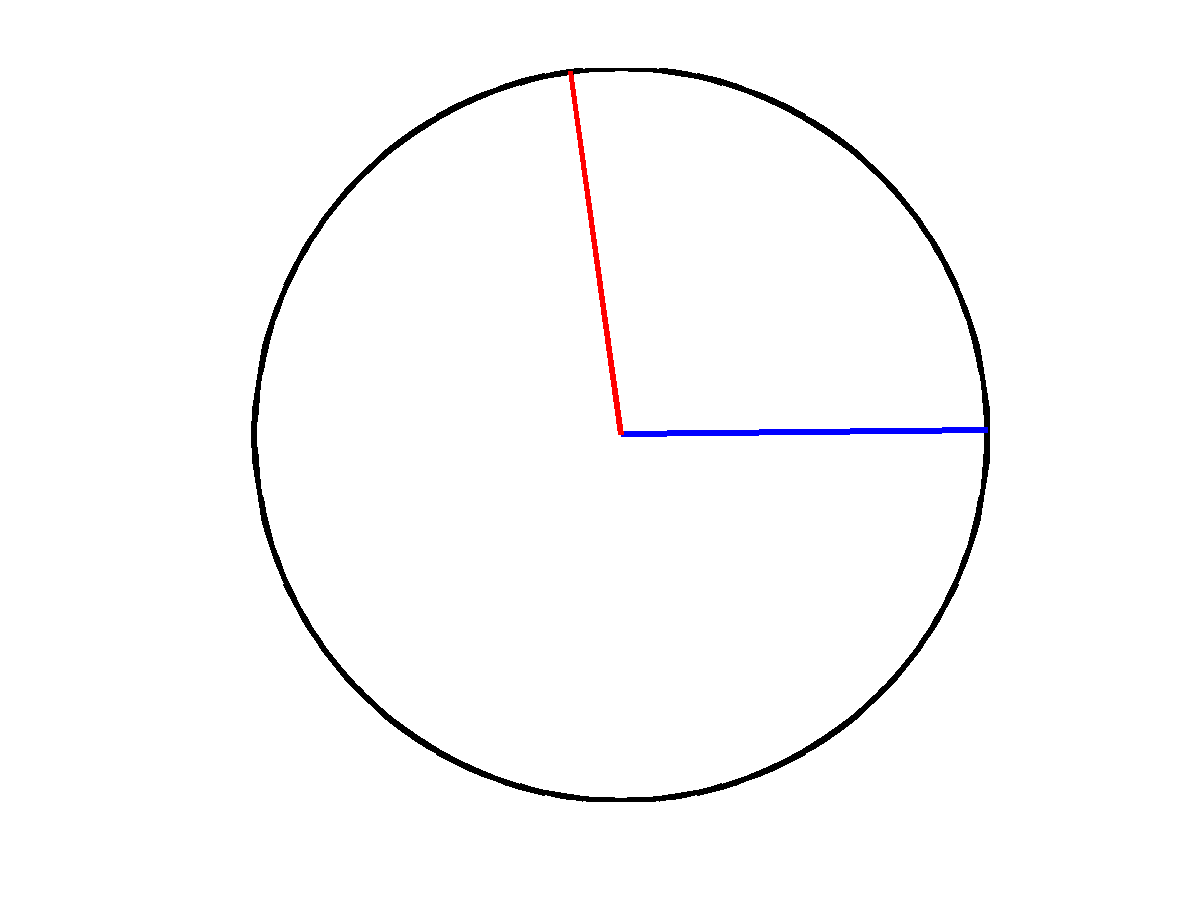} &
\includegraphics[width=0.51in]{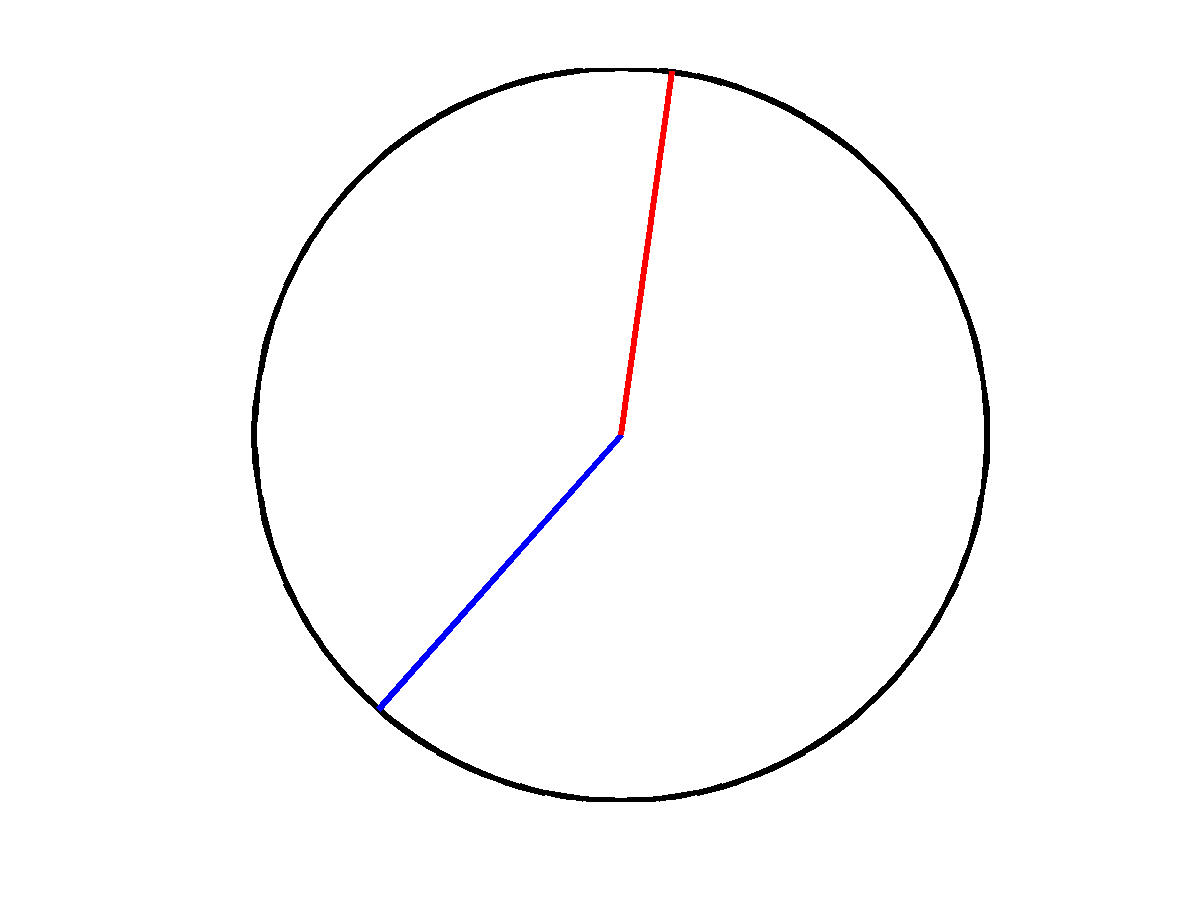} &
\includegraphics[width=0.51in]{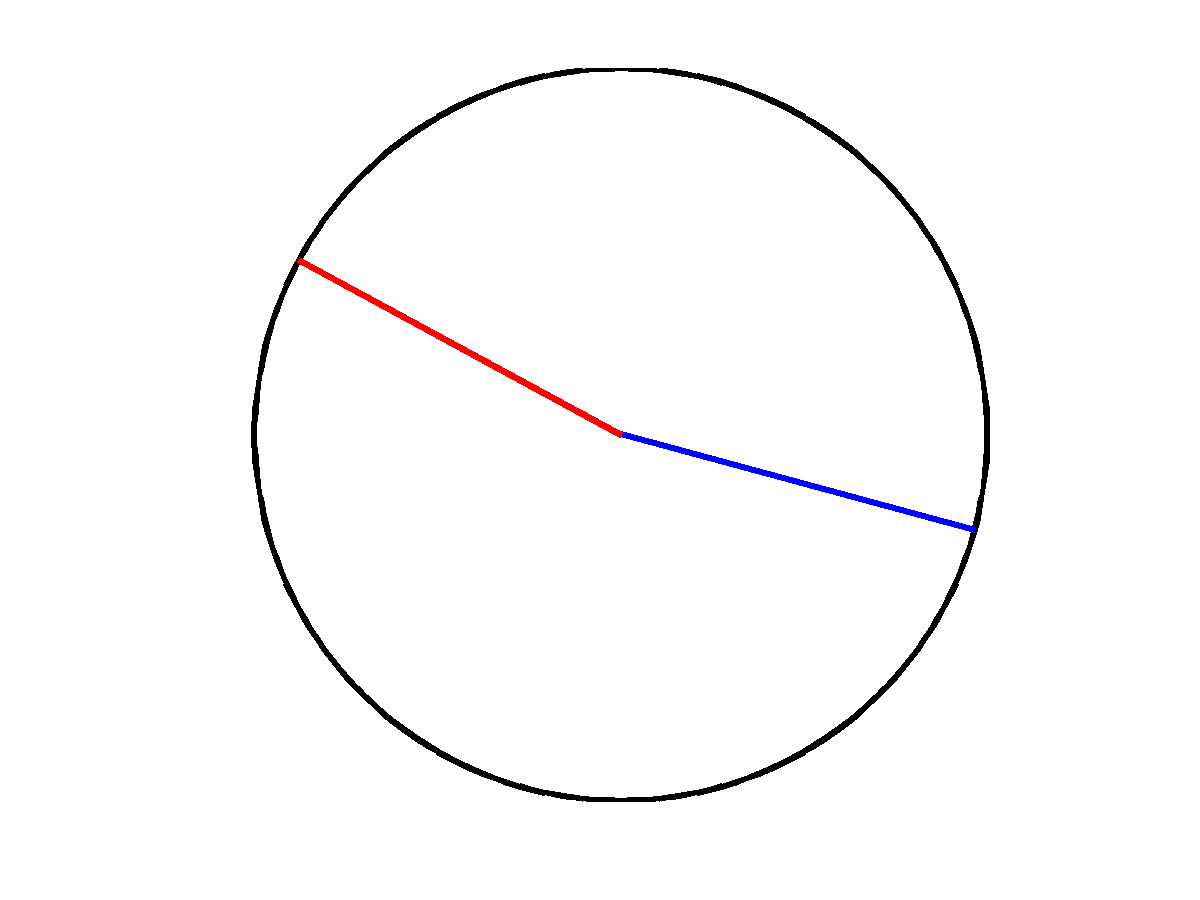} &
\includegraphics[width=0.51in]{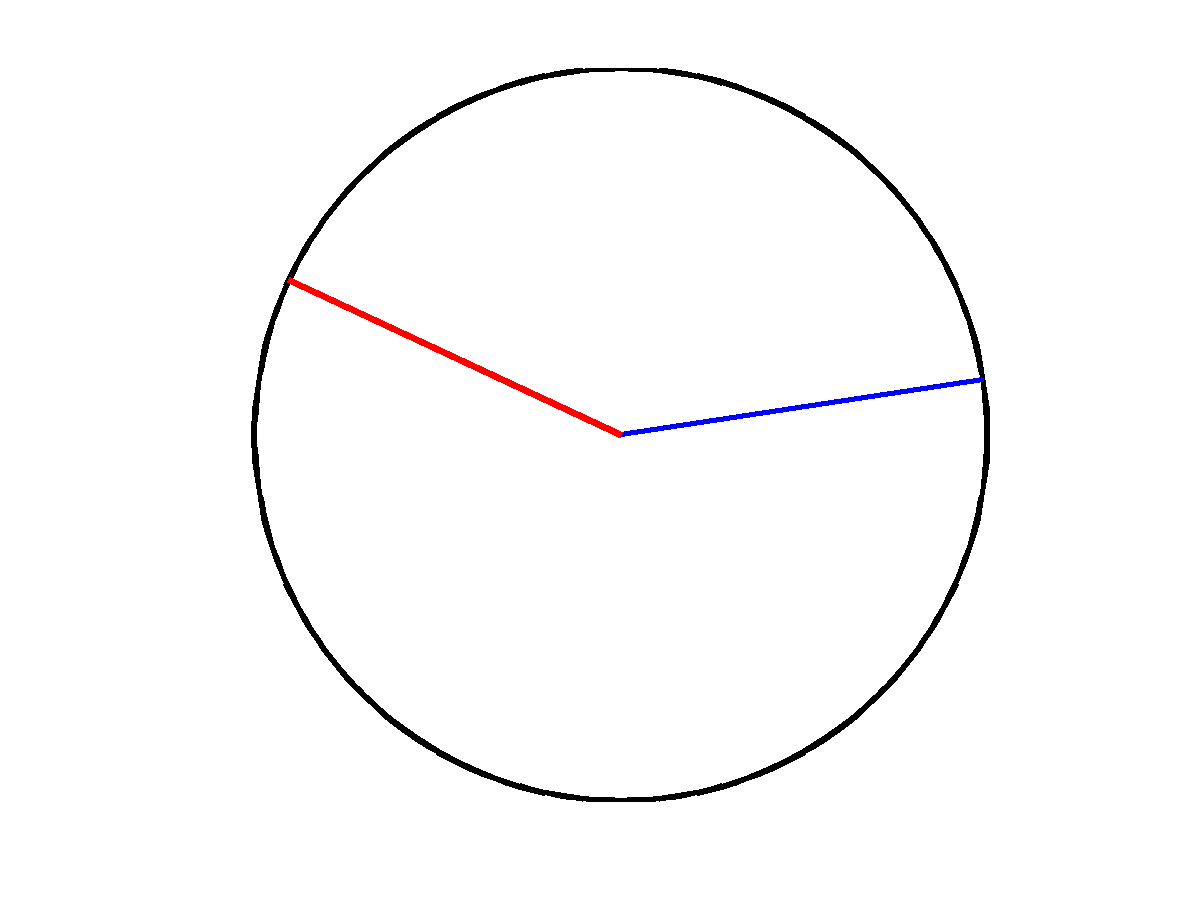} &
\includegraphics[width=0.51in]{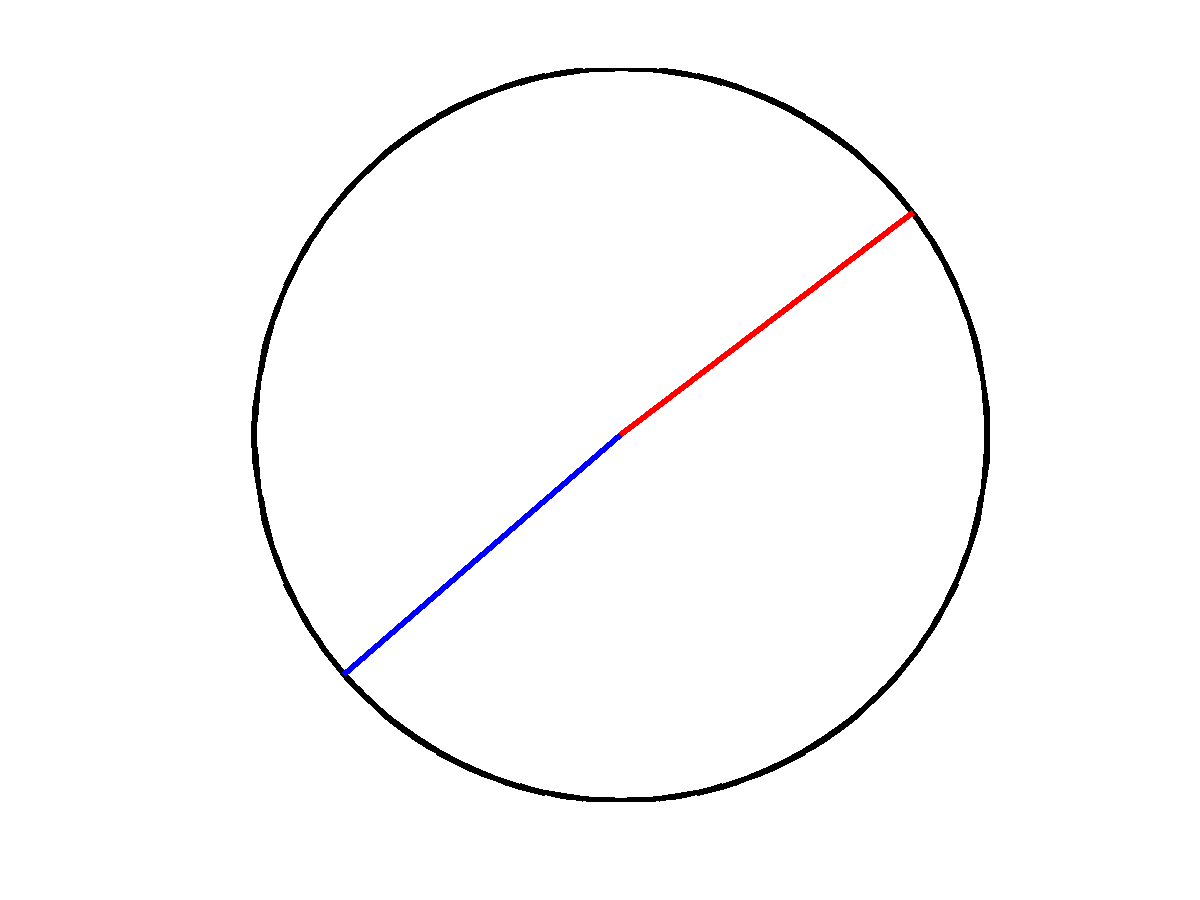}

\end{tabular}
\caption{Representative results obtained by the proposed method ( approach 3, $N=8,M=9$). A ground-truth orientation (red) and predicted orientation (blue) are indicated in a circle. The first three rows show successful cases while the last row shows failure cases.}\label{fig:visualResults_caffe}
\end{figure*}

Finally, Fig.~\ref{fig:visualResults_caffe} shows some representative results. The last row includes failure cases.

\begin{table*}[tb]
\small
\begin{center}
  \caption{Continuous pedestrian orientation estimation: Mean Absolute Error in degree, Accuracy-$22.5^\circ$ and Accuracy-$45^\circ$ are shown. \label{tb:TUD_results}}
  \begin{tabular}{|c|c|c|c|	} \hline
   Method & MeanAE ($^\circ$) & Accuracy-$22.5^\circ$ & Accuracy-$45^\circ$  \\ \hline
   \textbf{Ours} & 26.6 & 70.6 & 86.1 \\ \hline
   Hara and Chellappa \cite{Hara2016} & 34.7 & 68.6 & 78.0 \\ \hline \hline
   Human & 9.1 & 90.7 & 99.3 \cite{Hara2016} \\  \hline
   \end{tabular}
\end{center}
\end{table*}

\section{Conclusion}\label{sec:conclusions}
This work proposed a new approach for a continuous object orientation estimation task based on the DCNNs. Our best working approach works by first converting the continuous orientation estimation task into a set of non-overlapping discrete orientation estimation tasks and converting the discrete prediction to a continuous orientation by a mean-shift algorithm. Through experiments on a car orientation estimation task and a pedestrian orientation estimation task, we demonstrate that the DCNN equipped with the proposed orientation prediction unit works significantly better than the state of the approaches, providing another successful DCNN application. Our experiments also indicate that selecting a suitable representation is critical in transferring DCNNs trained on an image classification task to an orientation prediction task.
\newpage

{\bibliographystyle{ieee}
\bibliography{/Users/kotahara/papers/library}
}

\end{document}